\documentclass[twoside]{article}

\usepackage[accepted]{aistats2023}
\usepackage[round]{natbib}

\usepackage{enumitem}
\usepackage{float}
\usepackage{amsthm}
\newtheorem{theorem}{Theorem}
\newtheorem{lemma}{Lemma}

\newtheorem{assumption_2}{Assumption A.}
\newtheorem{theorem_2}{Theorem A.}

\usepackage{verbatim}
\usepackage{multirow}

\usepackage{xr}
\usepackage{balance}
\usepackage{xspace}

\usepackage{url}
\usepackage{amsmath}
\usepackage{amssymb}

\usepackage{mathtools}

\usepackage{dsfont}
\usepackage{upgreek}
\usepackage{bbm}			% mathbb vs mathds

\usepackage{graphicx}
\usepackage{epstopdf}
\graphicspath{ {./figures/} } % setups path to find graphics and images
\usepackage{grffile}
\usepackage{tikz}
\usepackage{float}

\usepackage{booktabs} % to thicken table lines (\toprule, \midrule, \bottomrule)

\usepackage[labelfont=bf]{caption}
\usepackage{subfigure}

\usepackage{mathtools}
\usepackage{relsize}  % for mathlarger to tune the size of math symbols

\newcommand{\Sec}[1]		{Sec.\,\ref{#1}}

\newcommand{\Fig}[1]		{Fig.\,\ref{#1}}

\newcommand{\Eq}[1]			{Eq.\,\ref{#1}}
\newcommand{\Expr}[1]			{Expr.\,\ref{#1}}

\newcommand{\Tab}[1]		{Tab.\,\ref{#1}}
\newcommand{\Alg}[1]		{Alg.\,\ref{#1}}
\newcommand{\Theorem}[1]{Theorem\,\ref{#1}}

\newcommand{\Lemma}[1]{Lemma\,\ref{#1}}

\newcommand{\Assumption}[1]{Assumption\,\ref{#1}}

\newcommand{\Problem}[1]{Problem\,\ref{#1}}	
\newcommand{\Appendix}[1]{Appendix\,\ref{#1}}	
\newcommand{\ie}   			{i.e.\@\xspace}
\newcommand{\eg}   			{e.g.\@\xspace}
\newcommand{\etc}   		{etc.\xspace}

\newcommand{\one}       {\mathbf{1}}     % or  {\mathbb{1}}
\newcommand{\ones}[1]   {\one_{#1}}
\newcommand{\zero}      {\mathbf{0}}
\newcommand{\zeros}[1]  {\zero_{#1}}
\newcommand{\diag}    {\operatorname{diag}}
\newcommand{\Block}    {\operatorname{Block}}
\newcommand{\real}      {\mathbb{R}}

\newcommand{\pdf}		    {pdf\xspace} %{p.d.f.\xspace}
\newcommand{\Hilbert}      {\mathbb{H}}

\newcommand{\PE}{P\!\!E}  % variables for Pearson's divergence

\newcommand{\fdiv} 	{$\phi$-divergence\xspace}
\newcommand{\fdivs} {$\phi$-divergences\xspace}
\newcommand{\KLdiv} {KL-divergence\xspace}
\newcommand{\PEdiv} {PE-divergence\xspace}
\newcommand{\PEdivs} {PE-divergences\xspace}
\newcommand{\pdfunc}		{pdf\xspace}
\newcommand{\pdfuncs}		{pdfs\xspace}
\newcommand{\inlinetitle}[2]  {\vspace{4pt}\noindent\textbf{\emph{#1}{#2}}}

\renewcommand*{\top}{{\mkern-1.5mu\mathsf{T}}}

\newcommand{\ExpecAlphaQv}[1]{\mathbb{E}_{p_v^{\alpha}(x)}[#1]}

\newcommand{\ExpecQm}[2]{\mathbb{E}_{q_{#2}(x)}[#1]}
\newcommand{\Hnull}{\text{H}_{\text{null}}} %{\text{H}_0}
\newcommand{\Halt}{\text{H}_{\text{alt}}} %{\text{H}_1} 
\newcommand{\norm}[1]{\left\lVert#1\right\rVert}

\newcommand{\dott}[2]{\langle #1,#2 \rangle}

\newcommand{\Lpc}{\mathlarger{\mathcal{L}}}
\newcommand{\id}{\mathbb{I}}

\newcommand{\abs}[1]{\left|#1\right|}

\newcommand{\setprev}{\mathbf{X}_v}
\newcommand{\setpostv}{{\mathbf{X}'_v}}

\newcommand{\X}{{\mathcal{X}}}
\newcommand{\nghood}{\text{ng}} % burning period
\newcommand{\Gdims}{N} % number of graph nodes
\newcommand{\G}{{\mathcal{G}}} % the graph symbol

\newcommand{\hthetanode}[1]{\hat{\theta}_{#1}}
\newcommand{\forwardtheta}[1]{\vec{\theta}_{#1}}

\newcommand{\forwardTheta}[1]{\vec{\Theta}_{#1}}
\newcommand{\backwardTheta}[1]{\cev{\Theta}_{#1}}

\newcommand{\forward}[1]{\vec{#1}}
\newcommand{\backward}[1]{\cev{#1}}

\newcommand{\q}{q}  %or p'

\newcommand{\SCORE}{S}
\newcommand{\eigmax}[1]{e_{\max}\!\left({#1}\right)}

\newcommand{\mydef}{=} %:=
\newcommand{\cev}[1]{\reflectbox{\ensuremath{\vec{\reflectbox{\ensuremath{#1}}}}}}

\newcommand{\likelihood}{likelihood\xspace}

\newcommand{\padded}[1] {\,#1\,}
\newcommand{\void}[1] {\padded{\cdot}}

\makeatletter
\newcommand{\pushright}[1]{\ifmeasuring@#1\else\omit\hfill$\displaystyle#1$\fi\ignorespaces}
\newcommand{\pushleft}[1]{\ifmeasuring@#1\else\omit$\displaystyle#1$\hfill\fi\ignorespaces}
\makeatother
\DeclareMathOperator*{\argmax}{\arg\!\max}
\DeclareMathOperator*{\argmin}{\arg\!\min}

\DeclareMathOperator*{\median}{median}

\DeclareMathOperator*{\R}{{\mathbb{R}}}

\usepackage[normalem]{ulem} % for strikethrough
\newcounter{marginNoteCounter}

\usepackage{algorithm}
\usepackage{algorithmic}
\usepackage{adjustbox}
\newcommand{\OCKG}{OCKG\xspace}

\usepackage{hyperref}

\begin{document}

\runningtitle{Online Centralized Non-parametric Change-point Detection}


\twocolumn[

\aistatstitle{Online Centralized Non-parametric Change-point Detection 
\\via Graph-based Likelihood-ratio Estimation}

\aistatsauthor{ Alejandro de la Concha  \And Argyris Kalogeratos  \And  Nicolas Vayatis }

\aistatsaddress{  Centre Borelli, ENS Paris-Saclay,\\
	Université Paris-Saclay, France   \\
  \texttt{{\smaller\{name.surname\}@ens-paris-saclay.fr} }} ]

\begin{abstract}
Consider each node of a graph to be generating a data stream that is synchronized and observed at near real-time. At a change-point $\tau$, a change occurs at a subset of nodes $C$, which affects the probability distribution of their associated node streams. %
In this paper, we propose a novel kernel-based method to both detect $\tau$ and localize $C$, based on the direct estimation of the likelihood-ratio between the post-change and the pre-change distributions of the node streams. Our main working hypothesis is the smoothness of the likelihood-ratio estimates over the graph, \ie connected nodes are expected to have similar likelihood-ratios. The quality of the proposed method is demonstrated on extensive experiments on synthetic scenarios.
\end{abstract}

\section{Introduction}\label{sec:intro}

Change-point detection is a fundamental problem in real time-series analysis and various control tasks. Modern challenges include handling larger amounts of more complex data streams, a clear example of which is when those lie over a graph. For example, many real-world systems can be seen as a network in which each node generates a stream of data: a network of seismic stations studying different geological events, the content shared by users of a social network, stations in a railway network, banks in a financial system, \etc 
A change-point may signify an earthquake, a shift of users' interest, a disruption of a railway service, or an early sign of an economic crisis. In these examples, the graph structure provides a priori relevant information about how the streams relate with each other, and maybe also shape their behavior after a change takes place. 

In this paper, we address a naturally arising question: 
how can we capitalize over the graph information in the online change-point detection task?

\inlinetitle{Related work}{.} Online, or sequential, change-point detection methods assume the user sees a stream of data in near real-time, and aim to detect the moment $\tau$ of a change-point as soon as possible, while minimizing the false alarm rate \citep{Tartakovsky2014,Tartakovsky2021,Xie2021}. Classical online change-point detection methods, such as Shewhart Chart \citep{Shewart1925}, CUSUM \citep{Page1954}, and Sriryaev-Roberts \citep{Shiryaev1963}, are based on the likelihood-ratio of the probability models related to the data streams before and after the change-point. For some parametric families, whose members admit probability density functions $p_{\theta}$ and $p_{\theta'}$ described by a set of parameters $\theta,\theta'$, the aforementioned methods assume they know these parameters and hence achieve an optimal trade-off between detection delay and false alarms \citep{Tartakovsky2021,Xie2021}. Nevertheless, the  hypothesis of complete knowledge of $p_{\theta}$ and 
$q_{\theta}$ is very restrictive in practice, mainly since the user rarely knows what to expect as system behavior after a change-point. 

In \cite{Nguyen2007}, the authors identified situations where the \fdiv estimation between two measures amounts to inferring the likelihood-ratio as an element of a functional space, which means doing so indirectly using data observations coming from $p$ and the assumed $\q$. That work motivated the development of non-parametric approaches for change-point detection based on approximations of the likelihood-ratio. The methods of this line of work have the advantage of being agnostic to both the probabilistic models generating the data stream and the characteristics of the possible change \citep{Yamada2011,Kanamori2012,Aminikhanghahi2017}. 

Both parametric and non-parametric methods were first applied to monitor a single data stream, but there has been a growing interest in extending these techniques to multiple streams. The latter refers to the case where each of the streams is associated with one node of a graph. Intuitively, the graph structure may carry relevant information for the fast detection of change-points. For the parametric case, we can point to \cite{Zou2018,Zou2019} where the knowledge of the process generating the streams is assumed, as well as that the expected change will affect a connected subset of nodes. For the non-parametric case, \cite{Ferrari2020} proposed a method where the underlying graph has community structure, and a change may occur in only one community (cluster of nodes). 

\inlinetitle{Contribution}{.} In this paper, we present the Online Centralized Kernel- and Graph-based (\OCKG) detection method, which is built upon a non-parametric likelihood-ratio estimation and the notion of graph smoothness. The latter concept formalizes the intuition that two nodes are expected to have a similar behavior before and after a change-point if they are connected. More precisely, our approach is built upon the graph-based likelihood-ratio estimation framework of \citep{delaConcha2022}. The \OCKG method %The aforementioned method 
has the notable advantages that it is: i) non-parametric and hence requiring minimum hypotheses about the nature of the data generating process at each of the graph nodes, ii) more sensitive compared to methods that aggregate all data streams in one stream, thanks to the integration of the graph structure, iii) more accurate in localizing the affected nodes, when compared with similar methods (\eg \cite{Ferrari2020}).

\section{Background and problem statement} \label{sec:setting}

\inlinetitle{General notations}{.} %\NEW{For reader's convenience, we summarize some notation used throughout the paper}. 
Let $a_i$ be the $i$-th entry of a vector $a$; when the vector is itself indexed by an index $j$, then we refer to its $i$-th entry  by $a_{j,i}$. Similarly, 
$A_{ij}$ is the entry at the $i$-th row and $j$-th column of a matrix $A$. $\eigmax{A}$
denotes the maximum eigenvalue of $A$.
$\ones{M}$ represents the vector with $M$ ones (resp. $\zeros{M}$), and $\id_{M}$ is the $M \times M$ identity matrix. %
We denote by $G=(V,E,W)$ a positive weighted and undirected graph% without self-loops
, where $V$ is the set of vertices, $E$ the set of edges, and $W \in \real^{\Gdims\times \Gdims}$ %denotes 
its %the associated 
adjacency matrix. The graph has no self-loops, \ie $W_{uu}=0, \forall u \in V$. The degree of $v$ is $d_v=\sum_{u \in \nghood(v)} W_{uv}$, where $u \in \nghood(v)$ indicates that $u$ is a neighbor of $v$. The degree of node $v$ is indicated by $d_v$; $\nghood(v) = \{u\,:\,W_{uv}\neq 0\}$ is the set of all the neighbors of $v$. With these elements, we can define the combinatorial Laplacian operator associated with $\G$ as $\Lpc=\diag((d_v)_{v \in V}) - W$, where $\diag(\cdot)$ is a diagonal matrix with the elements of the input vector in its diagonal.
 Finally, we will later find useful the notion of \emph{graph function}, which is any function $\omega:V \rightarrow \real^d$ assigning a vector in $\real^d$  to each node of a graph. When $d=1$, $\omega$ defines a \emph{graph signal} \citep{Shuman2013}.

\inlinetitle{\fdivs and likelihood-ratio}{.} %
\fdivs provide a way to measure the similarity between two probability measures. When both measures admit a \pdfunc, let those be $p$ and $\q$, respectively, in terms of the Lebesgue measure, then \fdiv is defined as:
$I_{\phi}(p,\q)=\int p(x)\, \phi\!\big(\frac{\q(x)}{p(x)}\big) dx$, for $x\in\X$.  
where  $\phi:\real \rightarrow \real$ is a convex and semi-continuous real function such that $\phi(1) = 0$ \citep{Csiszar1967}. Notably, for $\chi \in \real$, when $\phi(\chi)=\log(\chi)$ we recover the well-known Kullback–Leibler's (\KLdiv) \citep{Kullback59} which is omnipresent in most optimality theorems for online change-point detection methods \citep{Tartakovsky2014,Tartakovsky2021,Xie2021}. When $\phi(\chi)=\frac{(\chi-1)^2}{2}$, we identify the Pearson's \PEdiv \citep{Pearson1900}. 

The quantity $r(x)=\frac{q(x)}{p(x)}$ is called \emph{likelihood-ratio}, and is central in the computation of any \fdiv. As we will see in \Sec{sec:estimation}, we can translate
the approximation of a \PEdiv between $p$ and $q$ to a likelihood-ratio estimation problem. In practice, though, $r$ may be an unbounded function, challenging non-parametric methods that may fail to converge. %\NOTE{with the number of observations}. 
For this reason, a known workaround is to replace $p$ by $p^{\alpha}(x)=(1-\alpha)p(x)+\alpha q(x)$, and use instead the $\alpha$-\emph{relative \likelihood-ratio function} \citep{Yamada2011}: 
\begin{equation}\label{eq:rel-density-ratio}
%\begin{aligned}
r^{\alpha}(x) = \frac{q(x)}{(1-\alpha)p(x)+\alpha q(x)}
= \frac{q(x)}{p^{\alpha}(x)}
\leq \frac{1}{\alpha}, 
%\text{for any } 0 \leq \alpha < 1 \ \ \ \text{and} \ \ x \in \X,
%\end{aligned}
\end{equation}
for any $0 \leq \alpha < 1$, 
%and 
$x \in \X$%; 
%where 
%and $p^{\alpha}(x)=(1-\alpha)p(x)+\alpha q(x)$
.

\inlinetitle{Proposed setting and problem statement}{.} %
Let us suppose we observe $\Gdims$ synchronous data streams, each generated  by a node of a connected graph $G$. %We denote by 
Let $x_{v,t}$ be the observation %generated by 
at node $v$ at time $t$. We suppose that, $\forall v \in V$ and $t \in  \{1,...\}$, $x_{v,t} \in \X$ belongs to the same input space $\X \subset \R^d$. Furthermore, the observations are independent in time, which is a standard hypothesis in kernel-based change-point detection literature \citep{Arlot2019,Li2019,Harchaoui2008,Bouchikhi2019}.
 
\begin{figure}[t!]
%\begin{minipage}[b]{\linewidth}
  \centering
  %\centerline{
  %\includegraphics[width=0.9\linewidth, viewport=1 30 750 490, clip]{graph_signal_fixed.png}\hspace{4mm}%
  \includegraphics[width=\linewidth]{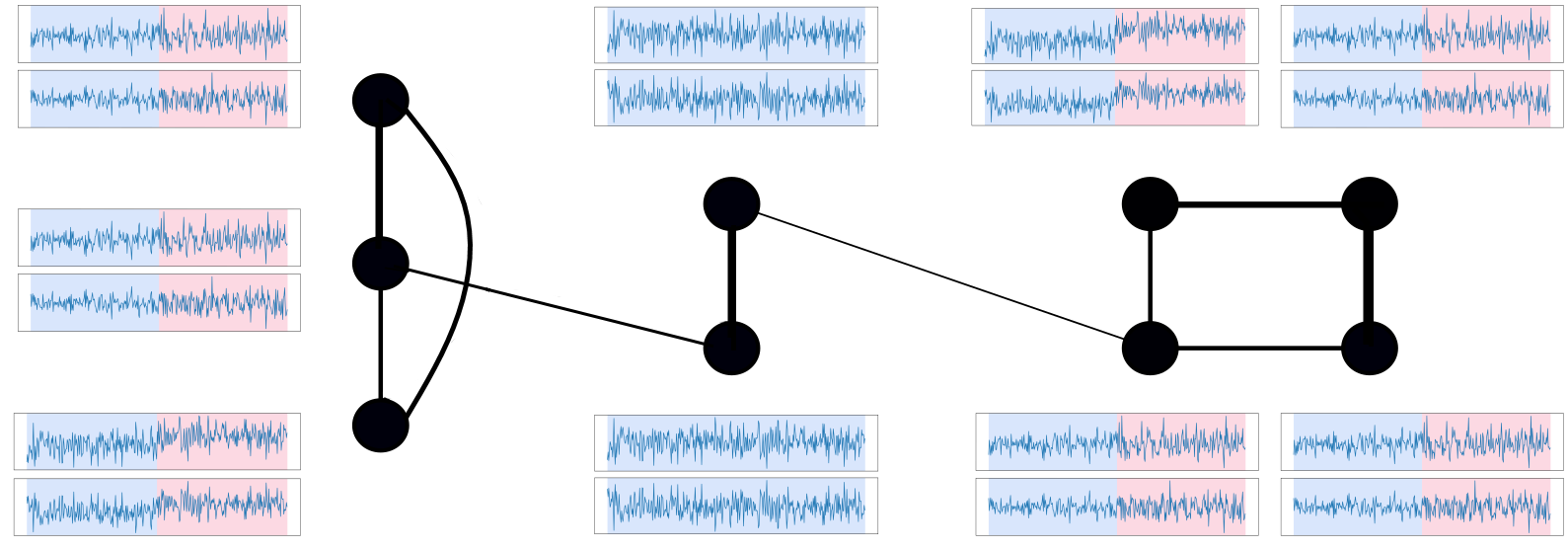}%
	%}
%\end{minipage}
\vspace{-0.5mm}
    \caption{\small Example of two-dimensional data streams observed over the nodes of a weighted graph. A change occurs in a subset of nodes, and the change-point is at the moment when the color changes %(from blue to pink) 
in the respective time-series. %As it can be seen, 
Left nodes: their change-point is associated with a shift in the covariance matrix. Right nodes: they experience a shift in the mean of their streams.}
\label{fig:CP_GS}
\end{figure}

Consider as change-point the timestamp $\tau$ at which the distribution associated with the streams observed at nodes belonging to a set $C$, changes:
\begin{equation}
    \left\{
    \begin{array}{ll}
        t < {\tau} \ \  x_{v,t} \sim p_v;  \\
        t \geq {\tau} \ \  x_{v,t} \sim q_v;
    \end{array}
    \right.
\end{equation}
where $p_v \neq q_v$ if ${v \in C}$, otherwise $p_v = q_v$. We consider all $p_v$, $q_v$, $C$, $\tau$ to be unknown. Moreover, we expect $C$ to depend on the graph structure. A simple example with signals $\X \subset \R^2$ at each node, is shown in \Fig{fig:CP_GS}.

For each node $v$, let us define the sample of $n$ consecutive observations, indexed by $t$, to be the set: 
\begin{equation}{\label{eq:windows}}
   \X_{v,t}= [x_{v,t-n},\,x_{v,t-(n-1)},\,...,\,x_{v,t-1}].
\end{equation}
Our method aims to compare the two adjacent samples, $\X_{v,t}$ and $\X_{v,t+n}$, for each node. More precisely, we learn jointly the $\Gdims$ relative likelihood-ratios $r^{\alpha}(\cdot)=(r_1^{\alpha}(\cdot),...,r_{\Gdims}^{\alpha}(\cdot))^\top \in \R^{\Gdims}$ (see \Eq{eq:rel-density-ratio}) and use them to design a score that indicates whether $\X_{v,t}$ and $\X_{v,t+n}$ follow the same distribution.  

In the text, we call \emph{null hypothesis} ($\Hnull$) %, $\Hnull$, 
the case with no change, \ie when $p_v=p'_v$, $\forall v \in V$, which opposes the \emph{alternative hypothesis} ($\Halt$) where a change does exist.%is denoted by $\Halt$.

\section{Problem formulation and solution}\label{sec:Problem_formulation}

The proposed Online Centralized Kernel- and Graph-based (\OCKG) change-point detection method capitalizes over the connection between the approximation of the Pearson's \PEdiv and the likelihood-ratio estimation. This technique allows us to 
compare the samples $\X_{v,t}$ and $\X_{v,t+n}$. By definition, the \PEdiv is a non-symmetric similarity measure, \ie generally $\PE(p,q) \neq \PE(q,p)$. For this reason, we generate approximations for both quantities.

In summary, the \OCKG method comprises three main 
tasks to which the next subsections are devoted: 
\begin{enumerate}
    \item[\textbf{1.}] \emph{\textbf{Estimation}}: When a new observation arrives at time $t$, we estimate the vector of relative likelihood-ratios $\forward{r^{\alpha}}_t(\cdot)=(\forward{r^{\alpha}}_{1,t}(\cdot),...,\forward{r^{\alpha}}_{\Gdims,t}(\cdot))$, between the samples of observations $\X_{t}$, $\X_{t+n}$, and for the reverse sample order $\backward{r^{\alpha}}_t(\cdot)$ with samples
    $\X_{t+n}$, $\X_{t}$.
    \item[\textbf{2.}] \emph{\textbf{Detection}}: The estimated
    likelihood-ratios $\forward{r^{\alpha}}_t(\cdot)$ and $\backward{r^{\alpha}}_t(\cdot)$ are used to approximate the respective \PEdivs, $\hat{\PE}{}^{\alpha}_v(\X_{t},\X_{t+n})$ and $\hat{\PE}{}^{\alpha}_v(\X_{t},\X_{t+n})$. Then, we define node scores, $\{S_v\}_{v \in V}$, based on the latter approximations. Finally, the node scores are aggregated into a global score indicating whether a change has occurred in the system. 
    \item[\textbf{3.}] \emph{\textbf{Identification}}: Once a change-point is spotted, we use the node scores $\{S_v\}_{v \in V}$ to identify the nodes at which the change occurred, thus belonging to $C$. 
\end{enumerate}

\subsection{Estimation}{\label{sec:estimation}}
%As explained in the previous section, 
The \OCKG method relies on the quantification of the difference between the probability models %described by the 
\pdfuncs $p_v$ and $q_v$ based just on two adjacent $n$-sized samples $\X_{v,t}$ and $\X_{v,t+n}$. This is done by exploiting the interplay between the \fdiv and the likelihood-ratios; more precisely, let us quote the following lemma (Lemma 2 in \cite{Nguyen2007}) referring to the \fdiv of $p_v$ and $q_v$: 
\begin{lemma}\label{lemma_Nguyen}
For any class of functions $\mathcal{F}$ mapping from $\X$ to $\R$
%\NEW{$\mathcal{F} : \X \rightarrow \R$}
, we have the lower bound for the \fdiv: 
\begin{equation}
I_{\phi}(p,q) \geq \sup_{f \in \mathcal{F}} \int \big[  f(x) q(x) dx - \phi^*(f(x))p(x) \big].
\end{equation}
Equality holds if and only if the subdifferential  $\partial  \phi\big(\!\frac{q(x)}{p(x)}\!\big)$ contains an element of $\mathcal{F}$, where $\phi^*$ is the conjugate dual function of $\phi$, \ie  $\phi^*(v) = \sup_{u \in \R} [uv - \phi(u)]$.
\end{lemma}
When we fix $\psi(\chi)=\frac{(\chi-1)^2}{2}$, we recover the \PEdiv, and according to the \Lemma{lemma_Nguyen} we %can 
write: 
\begin{equation}{\label{eq:div_ratio}}
\PE(p,q)  = \int  r(x) q(x) dx - \int \frac{r^2(x)}{2}  p(x) dx - \frac{1}{2},
\end{equation}
where $r(x)=\frac{q(x)}{p(x)}$ is the likelihood-ratio. 
We can use \Eq{eq:div_ratio} to translate the approximation of the \PEdiv between $p$ and $q$ to a likelihood-ratio estimation problem. As mentioned in \Sec{sec:setting}, in practice $r$ can be unbounded and %may 
prevent the convergence of non-parametric methods. We thus use instead the $\alpha$-\emph{relative \likelihood-ratio} of \Eq{eq:rel-density-ratio}.

Our graph-related objective is to estimate jointly the $N$ node-level $\alpha$-\emph{relative \likelihood-ratio functions}
$r^{\alpha}_v(x)$, $v \in V$, each one being associated with the node $v$'s \pdfuncs $q_v$ and $q_v'$. Our two main hypotheses are that: i) each $r^{\alpha}_v(x)$ is an element of a Reproducing Kernel Hilbert Space (RKHS), and that ii) the graph signal $r^{\alpha}(x)=(r^{\alpha}_1(x),...,r^{\alpha}_{\Gdims}(x))$ is expected to be also smooth after the change-point. Notice that, when $p = \q$, $r^{\alpha}(x)$ is the constant vector $\ones{M}$, which is obviously a perfectly smooth graph signal. 

\subsubsection{Cost function}{\label{subsec:estimation}}%
Let us introduce the RKHS $\Hilbert$ equipped with a reproducing kernel $K : \X \times \X \rightarrow \R$, with the associated inner-product $\dott{\cdot}{\cdot}_{\Hilbert}$ and feature map $\phi(\cdot): \X \rightarrow \Hilbert$. 
Let $f_v(\cdot)$ be the function approximating $r^{\alpha}_v(\cdot)=\frac{q_v(\cdot)}{(1-\alpha)p_v(\cdot)+ \alpha q_v(\cdot)}=\frac{q_v(\cdot)}{p_v^{\alpha}(\cdot)}$; we suppose $f_v(\cdot)$ is a linear model of the form $f_v(x)=\dott{\theta_v}{\phi(x)}_{\Hilbert}$, where $\theta_v \in \Hilbert$.  In practice, we will have access to $L$ elements of a global dictionary $D= \{x_1,..., x_L\}$ shared by all nodes. %Therefore, $\phi(\cdot)$ is a feature map of the vectorized form $\phi(x) \mydef (K(x,x_1),...,K(x,x_L))$, $\forall x \in \X$ and $\theta_v \in {\R}^{L}$.
Therefore, the vectorized form of the kernel feature map is $\phi(x) \mydef (K(x,x_1),...,K(x,x_L))$, $\forall x \in \X$ and $\theta_v \in {\R}^{L}$. The linear model can now be expressed as:
\begin{equation}\textstyle
    f_v(x)= \sum_{l=1}^{L} \theta_{v,l}  K(x,x_l).
\end{equation}
And by definition it holds:
\begin{equation}
  \abs{f_v(x)-f_u(x)} \leq \norm{\phi}_{\Hilbert} \norm{\theta_u-\theta_v}_{\Hilbert}.
\end{equation}
If the norm of the feature map $\norm{\phi}_{\Hilbert}$ is bounded for all $x$, then we can guarantee that $f_v(x)$ and $f_u(x)$ are close if the parameters $\theta_u$ and $\theta_v$ are close as well. This means we can induce smoothness to the graph signal $f(\cdot)=(f_1(\cdot),...,f_{\Gdims}(\cdot))$ in terms of their parameters $\{\theta_v\}_{v \in V}$. We will denote the concatenation of the parameters of interest as $\Theta=(\theta_1,...,\theta_{\Gdims})$.

With these elements, we can define the optimization \Problem{problem:cost_function} that is made of two components: the first term is a least square cost function aiming to approximate the relative likelihood-ratio at the node-level; the second term induces smoothness to the functions $f_u$ and $f_v$ by making 
their associated parameters, $\theta_{u}$ and $\theta_{v}$ of any two connected nodes $u$ and $v$, to be similar, while controlling the risk of overfitting via the penalization term $\norm{\theta_u}^2$ \citep{Sheldon2008}:
\begin{equation}\label{problem:cost_function}
%\hspace{-4mm}
\begin{aligned}
&\min_{\Theta \in \real^{\Gdims L}} \frac{1}{\Gdims} \sum_{v \in V}\!  \frac{\ExpecAlphaQv{(r^\alpha_v(x)-f_v(x))^2}}{2} \\
&\quad\qquad+ \frac{\lambda}{2} \sum_{u,v \in V\!\!\!\!\!\!} W_{uv} \norm{\theta_{u}-\theta_{v}}^2+ \frac{\lambda \gamma}{2} \sum_{v \in V} \norm{\theta_v}^2 = \\
&\min_{\Theta \in \real^{\Gdims L}}  \frac{1}{\Gdims}  \sum_{v \in V}  \!\!\bigg( \frac{\ExpecAlphaQv{\dott{\theta_v}{\phi(x)}^2}}{2} -  \ExpecQm{\dott{\theta_v}{\phi(x)}}{v} \!\bigg) \\ 
&\quad\qquad+ \frac{\lambda}{2} \sum_{u,v \in V\!\!\!\!\!\!\!\!} W_{uv} \norm{\theta_{u}-\theta_{v}}^2 + \frac{\lambda \gamma}{2} \sum_{v \in V} \norm{\theta_{v}}^2 + \Xi.\!\!\!\!\!\!\!
\end{aligned}    
\end{equation}
The equality between the two expressions comes from the development of the square in the first term, and the fact $E_{q_{v}^{\alpha}}[(g(x)r^\alpha_v(x))]=E_{q'(v)}[g(x)]$. Note that $\gamma,\lambda>0$ are penalization  constants, and $\Xi$ is a term we can ignore since it does not depend on $\Theta$.

Let us suppose that, at time $t$, for each node $v$ we have access to observations coming from the two probabilistic models described by $q_v$ and $q'_v$, which we respectively denote by $\X_{v}$ and $\X'_{v}$. We define the elements: 
\begin{equation*}%{\label{eq:updatehs}}
\begin{aligned}
& \ \ H_{v,t} = \frac{1}{n}\sum_{x \in \X_{v}} \phi(x)\phi(x)^\top, \ \ \ \ 
H'_{v,t} = \frac{1}{n}\sum_{x \in \X'_{v}} \phi(x)\phi(x)^\top, %,\\
%& h'_{v,t} = \frac{1}{n}\sum_{x \in \X'_{v}} \phi(x).
\end{aligned}
\end{equation*}
\begin{equation}{\label{eq:updatehs}}
\begin{aligned}
& h'_{v,t} = \frac{1}{n}\sum_{x \in \X'_{v}} \phi(x).\qquad \qquad\qquad\qquad\qquad\qquad
\end{aligned}
\end{equation}

Then, by rewriting \Problem{problem:cost_function} in terms of empirical expectations, we get: 
\begin{equation}{\label{eq:thetas_cost_function}}
\begin{aligned}
& \min_{\Theta \in \real^{\Gdims L }}   \frac{1}{\Gdims} \sum_{v \in V }  \left( (1-\alpha) \frac{\theta_v ^\top  H_{v,t} \theta_v}{2} + \alpha  \frac{\theta_v ^\top  H'_{v,t} \theta_v}{2}  -  h'_{v,t} \theta_v \right) \\  
&\quad\qquad + \frac{\lambda}{4} \sum_{u,v \in V}  W_{uv} \norm{\theta_u-\theta_v}^2 + \frac{\lambda \gamma}{2} \sum_{v \in V } \norm{\theta_v}^2 \\
&= \frac{1}{\Gdims} \bigg(\sum_{v \in V} \ell_{v,t}(\theta_v)\bigg)+  \frac{\lambda}{2} \Theta^\top \left([\Lpc+\gamma \id_{\Gdims}] \otimes \id_{L}\right) \Theta, 
\end{aligned}
\end{equation}

where $\ell_{v,t}(\theta_v) \mydef (1-\alpha) \frac{\theta_v ^\top  H_{v,t} \theta_v}{2} + \alpha  \frac{\theta_v ^\top  H'_{v,t} \theta_v}{2}  -  h'_{v,t} \theta_v $, and $\otimes$ is the Kronecker product between two matrices.

Let us call $\hat{\Theta}_t=(\hat{\theta}_1^{\top},\hat{\theta}_2^{\top},...,\hat{\theta}_{\Gdims}^{\top})$ the solution of \Problem{problem:cost_function}, and $\{\hat{f}_{v,t}(\cdot)\}_{v \in V}$ the estimated relative likelihood-ratio. Then we can approximate the \PEdiv $\PE(q_v^{\alpha},q_v)$ using an empirical approximation of \Eq{eq:div_ratio} and the estimation $\hat{f}_{v,t}(\cdot)$:
\begin{equation}{\label{eq:pearson_div_approx}}
\begin{aligned}
& \hat{\PE}{}^{\alpha}_v(\X_{v,t},\X'_{v,t}) = 
\sum_{x' \in \setpostv} \frac{\hat{f}_{v,t}(x)}{n} 
- \frac{(1-\alpha)}{2} \left( \sum_{x \in \setprev} \frac{\hat{f}_{v,t}(x)^2}{n} \right)
\\  &\quad\  - \frac{\alpha}{2} \left(  \sum_{x' \in \setpostv} \frac{\hat{f}_{v,t}(x)^2}{n}\right)- \frac{1}{2}
\\  &= - \bigg(\!(1-\alpha) \frac{\hat{\theta}_{v,t}^\top  H_{v,t} \hat{\theta}_{v,t}}{2} + \alpha  \frac{\hat{\theta}_{v,t}^\top  H'_{v,t} \hat{\theta}_{v,t}}{2}  - h'_{v,t} \hat{\theta}_{v,t} \!\bigg) - \frac{1}{2} \ \ 
\\ & =  -\ell_{v,t}(\hat{\theta}_{v,t}) - \frac{1}{2}.
    \end{aligned}
\end{equation}
The lack of symmetry of \PEdiv is relevant in the change-point detection task, since at every time $t$ we need to compare the two samples associated with the probabilistic models described by $\{p_v\}_{v \in V}$ and $\{p'_v\}_{v \in V}$ respectively. Depending on which \pdf is taken as numerator in 
$r_v^{\alpha}(\cdot)$, the associated \PEdiv may vary its sensibility to detect change-points. For this reason, we estimate two parameters $\forwardTheta{t}$ and $\backwardTheta{t}$ to approximate 
 $\hat{\PE}{}^{\alpha}_v(\X_{v,t},\X_{v,t+n})$ and $\hat{\PE}{}^{\alpha}_v(\X_{v,t+n},\X_{v,t})$, respectively.

\subsubsection{Optimization procedure}{\label{subsec:optimization_procedure}}
\begin{equation}
\begin{aligned}
 H_t   & = \text{Block}(H_{1,t},...,H_{\Gdims,t}) \in \real^{ \Gdims L \times \Gdims L}, \\ H'_{t}  & = \text{Block}(H'_{1,t},...,H'_{\Gdims,t}) \in \real^{ \Gdims L \times \Gdims L}, \\
 h'_{t}  & = (h_{1,t}'^{T},...,h_{\Gdims,t}'^T) \in \real^{\Gdims L}.
\end{aligned}
\end{equation}
 With these terms, we can easily verify that the optimization \Problem{eq:thetas_cost_function} is in fact a Quadratic problem:
\begin{equation}\label{eq:final_problem}
\begin{aligned}
    &\min_{\Theta \in \R^{\Gdims L} } \Phi_t(\Theta) = \min_{\Theta \in \R^{\Gdims L} } \frac{ \Theta^\top (A_t)\Theta}{2}- \Theta^\top b_t \\
   & =
    \min_{\Theta \in \R^{\Gdims L} } \frac{ \Theta^\top ( \frac{1-\alpha}{\Gdims}H_t+\frac{\alpha}{\Gdims}H'_{t} + \left([\Lpc+\gamma \id_{\Gdims}] \otimes \id_{L} \right))\Theta}{2}  - \Theta^\top h'_{t}.
\end{aligned}
\end{equation}
Note that $A_t= \frac{1-\alpha}{\Gdims}H_t+\frac{\alpha}{\Gdims}H'_{t} + [\Lpc+\gamma \id_{\Gdims}] \otimes \id_{L}$ defines a positive definite matrix. Then, at every time $t$ we could obtain a closed-form solution $\hat{\Theta}_t= A_t^{-1} b_t$. Nevertheless, in real applications, the size of the graph and the dictionary may cause this matrix inversion to be infeasible. For this reason, we propose to solve the problem via the Cyclic Block Coordinate Gradient Descent (CBCGD) method \citep{Beck2013,Li2018}. 

In our formulation, each block of variables is associated with a node $v$, thus it contains $\theta_v$. Before applying CBCGD, we need to define a fixed order for update of the variables. Let $\theta_{<v}$ be the set of variables that were updated before $v$, and $\theta_{ \geq v}$ be the complement of that set. Then, the $i$-th update of the parameter $\hat{\theta}_{v,t}$ is performed according to the schema: 
\begin{equation}\label{eq:update}
\begin{aligned} 
\!\!\!\!\!\!\!\hthetanode{v,t}^{(i)} &  \mydef  \frac{1}{\eta_{v,t} + \lambda \gamma} \Bigg[ 
\eta_{v,t} \hthetanode{v,t}^{(i-1)}  \\ 
& - \overbrace{
 \bigg(\!\frac{(1-\alpha) H_{v,t} + \alpha H'_{v,t}}{\Gdims} \hthetanode{v,t}^{(i-1)} - \frac{h'_{v,t}}{\Gdims} \!\bigg)}^{\mathclap{\text{component depending on node } v}} \\
  &- \overbrace{\lambda \bigg(\! d_v \hthetanode{v,t}^{(i-1)}- \!\!\!\sum_{u \in \nghood(v)\!\!\!\!\!\!\!\!} W_{uv} \big( \hthetanode{u,t}^{(i)} \one_{u<v} + \hthetanode{u,t}^{(i-1)} \one_{u \geq v} \!\big) \!\bigg)\!\!}^{\mathclap{\text{component depending on the graph}}} \Bigg]\!.
\end{aligned}
\end{equation}
An important point to highlight is the behavior of this schema when the graph structure is ignored. If we set $W = \zeros{M \times M}$, we recover a gradient descent algorithm running independently for each node (\ie ignoring the graph). This variant is called Pool in the experiments of \Sec{sec:exps}.

Notice that when no change has occurred, we expect the problem instances $\Phi_t$ and $\Phi_{t+1}$ to be similar. Thus, in this case, we can use the solution $\hat{\Theta}_{t-1}$ to initialize the problem for solving $\Phi_{t+1}$. More over, we can prove that the fact our problem is quadratic the number of cycles required to generate a new estimate $\hat{\Theta}_t$ from a new observation arriving to all the nodes and the previous estimate $\hat{\Theta}_{t-1}$ scales nicely even in big graphs. This statement can be found in \Appendix{appendix:optimization} as \Theorem{Th:convergence}. The conclusion of the theorem is, first, that the number of cycles  between points from time $t$ to time $t+1$ remain manageable, due to the efficient initialization of $\Phi_{t+1}$ with $\Theta_{t-1}$ (as explained earlier); 
second, the graph size impacts the optimization at a rate of $O(\log^2(N))$, which is manageable for many real graphs.  

\subsection{Detection and identification}
A well-known property of $\PE(p,q) \geq 0$, is that it becomes zero if and only if $p=q$, which makes it a good candidate as a score to validate whether a change exists \citep{Kawahara2012}. Then, the definition of a node-level score comes naturally:
\begin{equation}{\label{eq:node_scores}}
S_{v,t}=\max \{\hat{\PE}{}^{\alpha}_v(\X_{v,t},\X_{v,t+n})+\hat{\PE}{}^{\alpha}_v(\X_{v,t},\X_{v,t+n}),0 \}.
\end{equation}
The maximum is taken as the approximations can be negative. 
Next we define the global score $\SCORE_t=\sum_{v \in V} S_{v,t}$, which triggers a global alarm when $\SCORE_t \geq \eta$, where $\eta>0$ is a threshold parameter fixed by the user. The moment $\tau$ at which this global alarm fires, is also the estimated occurrence time of the associated change-point.

Once a change-point has been detected, we need to identify the affected nodes of the subset $C$. We select the nodes that satisfy $S_{v,t}>\eta_v$, where $\{\eta_v\}_{v \in V}$ is a set of positive constants given by the user.

The detailed pseudocode of \OCKG is described in \Alg{alg:OCKGD}.  Notice that by design, we expect $\hat{\PE}_v(\cdot,\cdot)$ to achieve its maximum value when it compares $\X_{v,\tau}$ and $\X_{v,\tau+n}$, which means there is always a lag of size $n$ (observations) in the detection of $\tau$. We desire $n$ to be as small as possible, yet guarantying a good identification of the nodes of interest. We further discuss this point in the experiments of \Sec{sec:exps}.
%%%%
\subsection{Further implementation details}{\label{sec:practical_implementation}}
%%%%
\inlinetitle{Dictionary}{.}{\label{subsec:dl}} In the previous section, we have made the implicit  hypothesis that we have access to a predefined dictionary $D$ and that it remains stable through time. This can be hard in practice, especially since in a change-point detection task we expect that at some point the observations will start being different to what will have been seen till then. Therefore, it is important to update the dictionary with new observations in an online manner, in order to improve the quality of the estimators $\hat{f}_{v,t}(\cdot)$. More specifically, we follow the approach that was suggested by \cite{Cedric2009} in the context of online prediction for time-series, where new observations get included into a dictionary, while controlling its size and the sparse representation of the datapoints. This is achieved by the \emph{coherence} measure that quantifies the redundancy between the dictionary elements as being linearly dependent. If this is smaller than a given threshold $\mu_0$, the new datapoint is added into the dictionary. After reaching a maximum dictionary size $L$, we delete the datapoint with the highest coherence (\ie highest redundancy) each time we insert a new element to the dictionary.

\inlinetitle{Hyperparameter Selection}{.}{\label{subsec:hyperparematers}} An important question that remains to be answered is how to select the hyperparameters associated with the kernel $K$, and the penalization constants $\lambda$ and $\gamma$. As in previous works in non-parametric \fdiv estimation, we use a cross-validation strategy \citep{Sugiyama2007,Sugiyama2011,Yamada2011}. This tries to minimize the sum of the losses $\frac{1}{\Gdims} \sum_{v \in V} \ell_{v,t}(\theta_v)$, which is equivalent to maximizing the mean node-level {\ignorespaces\PEdiv}s. The hyperparameter selection procedure is described by \Alg{alg:model_selection} included in \Appendix{appendix:hyper_selection}. In order to keep the computational cost reasonable, we recommend estimating the right hyperparameters just once in each direction. In the experiments, we applied this approach using a set of observations that had no change-points. 

The value of $\alpha$ is important for the stability of the algorithm. In the univariate case, it has been shown that the convergence rates of the cost function inspired by the Pearson Divergence depends on $\norm{r_{\alpha}}_{\infty}$ \cite{Yamada2011}. This means that a higher $\alpha$ value is expected to improve the convergence with respect to the number of observations, but on the other hand it will make harder to identify whether $p$ and $q$ differ. The role of $\alpha$ value is also investigated in our experiments.
%%%
\begin{algorithm}
\small
%\SetAlgoLined
%\SetAlgoLined
   \caption{The OCKG detector \!\!\!\!}\label{alg:OCKGD}
\begin{algorithmic}[1]
\STATE \textbf{Input:}\\  
$\alpha \in [0,1)$ : parameter of the relative \likelihood-ratio (%see
\Eq{eq:rel-density-ratio});\\
$n$: the size of the sample to use; \\
$D_1,D_2$: precomputed dictionaries, with $L_1$ and $L_2$ elements, respectively; \\ 
$(\sigma^*_1,\lambda_1^*,\gamma_1^*),(\sigma^*_2,\lambda_2^*,\gamma_2^*)$: optimal hyperparemeters (%see
\Alg{alg:model_selection});
\\
$\mu_0,L:$ coherence threshold controlling the dictionary updates, and the maximum dictionary size; \\
$tol:$ tolerated relative error for the optimization process; \\
$\eta, \{\eta_v\}:$ threshold to raise a global alarm, $\eta_v$ threshold to raise an alarm at node $v$. \\
\STATE \textbf{Output:} $\hat{\tau}$: the detection time of the change;\\ $\hat{C}$: the set of nodes where the change is observed. \\
\vspace{1mm}
\hrule
\vspace{1mm}
\raisebox{0.25em}{{\scriptsize$_\blacksquare$}}~\textbf{Initialization of parameters}

\STATE  $\forwardTheta{n}^{(0)}= 
\backwardTheta{n}^{(0)}=\zeros{L\Gdims}$

\raisebox{0.25em}{{\scriptsize$_\blacksquare$}}~\textbf{Online estimation and detection}

\FOR{$t \in \{n,...,\}$} 

\FOR{$v \in \{1,...,\Gdims\}$} 
\STATE Observe $x_{v,t+n-1}$ and update the sliding windows $\X_t,\X_{t+n}$ (\Eq{eq:windows})\!\!\!\!\!\!\!\!\!\!

\raisebox{0.25em}{{\scriptsize$_\square$}}~\textbf{Dictionary update}

\IF{$\max_{l \in \{1,...,L_1 \}} k(x_{v,l},x_{v,t+n-1}) \leq \mu_0$}

\STATE Add $x_{v,t+n-1}$ to the dictionary $D_1$
\STATE if the maximum dictionary size is reached, delete the datapoint with the highest coherence 
\ENDIF 

\ENDFOR

\raisebox{0.25em}{{\scriptsize$_\square$}}~\textbf{Parameters update}

\STATE Define $\vartheta_{v} = [\theta_{v,t-1}^{\top}, \zeros{d_1}]$, ($d_1$ is the number of new elements added to the dictionary)
\STATE Initialize $\forwardtheta{v,t-1}^{(0)}=\vartheta_{v}$

\STATE Fix $\X=\X_{t}$ and $\X'=\X_{t+n}$

\FOR{$v \in \{1,...,\Gdims\}$} 

\STATE Compute the quantities $H_{v},H'_{v},h'_{v}$. \hfill(see \Eq{eq:updatehs})

\STATE Fix  $\eta_{v,t}= \eigmax{\!\frac{ (1-\alpha) H_{v,t} + \alpha H'_{v,t} }{\Gdims} + \lambda d_v \id_{L_1} \!}$ 

\ENDFOR

\WHILE{$\norm{\forwardTheta{t}^{(i)}-\forwardTheta{t}^{(i-1)}}>\epsilon$} 
\FOR{$v \in \{1,...,\Gdims\}$} 

\STATE Update $\forwardtheta{v}^{(i)}$ \hfill(see \Eq{eq:update})

\ENDFOR
\ENDWHILE

\FOR{$v \in \{1,...,\Gdims\}$}

\STATE  Estimate  
$\hat{\PE}{}^{\alpha}_v(\X_{v},\X'_{v})$ \hfill(see \Eq{eq:pearson_div_approx})
\ENDFOR

\STATE Fix $\X=\X_{t+n}$ and $\X'=\X_{t}$

\STATE Repeat steps $6$\,--\,$22$ to compute $\backwardTheta{t}$ and $\hat{\PE}{}^{\alpha}_v(\X'_{v,t},\X_{v,t})$

\STATE \raisebox{0.25em}{{\scriptsize$_\square$}}~\textbf{Online detection and Identification} 

\STATE Compute the node scores $\SCORE_{v,t}$ \hfill(see \Eq{eq:node_scores})

\STATE Compute the global score $\SCORE_{t}=\sum_{v \in V} \SCORE_{v,t}$ 

\IF{ $\SCORE_t > \eta$}

\STATE A change-point is detected at $\hat{\tau}=t$

\IF{ $\SCORE_{v,t} > \eta_{v}$} 
\STATE Add $v$ to $\hat{C}$
\ENDIF
\ENDIF
\ENDFOR
 
\STATE Return $\hat{\tau}$ and $\hat{C}$

\end{algorithmic}
\end{algorithm}
%%%

\section{Experiments}\label{sec:exps}

\inlinetitle{Competitors}{.}
In this section we compare the performance of the \OCKG detector against two other alternatives that are based on non-parametric estimation. We refer to those methods as Nougat and Pool in the reported results.

Nougat is a non-parametric method aiming to detect a change in a cluster of graph nodes \citep{Ferrari2020,Ferrari2021}. It estimates the node-level likelihood-ratio via kernel methods and a stochastic gradient descent. At every time $t$ a single step of stochastic gradient descend is performed, and the updated function is then evaluated at time $t+1$ with the new incoming observation. This is done independent for each of the nodes of the graph. The resulting evaluation of the estimated function is used to construct a graph signal. Finally, Nougat filters the signal with the Graph Fourier Scan Statistic (GFSS), a graph-based statistical test that has been used for detecting nodes with anomalous activity in a graph \citep{Sharpnack2016}. The score for node $v$ is the absolute value of the $v$-th entry of the filtered signal, and the global score is the norm of this vector. 

Pool is a variant of the proposed \OCKG that ignores the graph structure, \ie when we set $W=\zeros{M \times M}$, and thus serves a baseline for investigating the benefits of using the graph. In this configuration, there is a global dictionary available for all the nodes, which is updated as observations become available, and the updates of the node parameters $\theta_v$ are carried out independently for each of the nodes. This detector can be seen as a RULSIF-based adaptation \citep{Sugiyama2007}. RULSIF is a state-of-the-art approach that has proved to be superior empirically and theoretically, when compared with other likelihood-ratio based methods \citep{Kawahara2012}. 

\inlinetitle{Setup}{.} 
For all the detectors, we focus on the Gaussian kernel, which means that among the hyperparameters to optimize there are the associated $\sigma$ parameters of this type of kernel.  
Then, for each compared detector, there is a different set of regularization parameters to tune. For \OCKG the regularization terms are $\lambda$ and $\gamma$, while just  $\gamma$ for Pool and Nougat. For Pool and \OCKG, we run the method described in  \Alg{alg:model_selection} in a set of observations of size $2n$ where there are no change-points. The same set of observations is used to build the initial dictionary according to the approach described in \citep{Cedric2009} with the coherence threshold parameter $\mu_0=0.1$. The same coherence threshold is used for the online updating of the dictionary. 

For Nougat, the selection of the hyperparameters $\sigma$ and $\lambda$ is not easy, as this was left as an open question in the original papers presenting this approach. To overcome this limitation we run the same procedure that we apply to Pool, but using $\alpha=0$, since Nougat in fact estimates the unregularized likelihood-ratios $r_v(x) - 1 =\frac{q_v(x)}{p_v(x)} - 1$. On the other hand, for the dictionary building strategy, Nougat referred to \citep{Ferrari2021} that uses a similar approach as the one we use for \OCKG and Pool. There are two more parameters to be fixed in Nougat: the learning rate $\eta_v$ of the gradient descent, and a parameter associated with the GFSS and the spectrum of the graph. For the learning rate, we fix $\eta_v=\frac{1}{10 \eigmax{(H_{v,0} +\gamma \id_L)}}$ inspired by the convergence guarantees provided in \citep{Ferrari2021}, where $H_{v,0}$ is build with $n$ observations coming from $p_v(\cdot)$ (\Eq{eq:updatehs}). Finally, we pick the smallest non-zero eigenvalue of the Laplacian as the tuning parameter of the GFSS filter, which also provided the best results. 

In order to select the width $\sigma$ for the Gaussian kernel, we first compute $\{\sigma_{v}\}_{v \in V}$ for each node via the median heuristic applied to the observations of $X_v$ (such quantities are available when generating the dictionary), and we define $\sigma_{\text{min}}= \argmin \{\sigma_{v}\}_{v \in V}$, $\sigma_\text{median}= \median\{\sigma_{v}\}_{v \in V}$ and $\sigma_{\text{max}}= \argmax \{\sigma_{v}\}_{v \in V}$, we then chose the final parameter from the set $\{\sigma_{\text{min}},\frac{1}{2}(\sigma_{\text{min}}+\sigma_\text{median}),\sigma_\text{median},\frac{1}{2}(\sigma_{\text{max}}+\sigma_\text{median}),\sigma_{\text{max}}\}$. $\gamma$ is selected from the set $\{1e^{-5},1e^{-3},0.1,1\}$. Finally, we define the average node degree $\bar{d}$, %=\frac{\sum_{v \in V} d_v}{\Gdims}$ 
and we identify the optimal $\lambda^*$ from the set $\{1e^{-3}\cdot\frac{1}{\bar{d}},1e^{-2}\cdot\frac{1}{\bar{d}},0.1\cdot\frac{1}{\bar{d}},1\cdot\frac{1}{\bar{d}},10\cdot\frac{1}{\bar{d}}\}$.

\subsection{Use-cases on synthetic data}
In this section, we present two different synthetic scenarios to test the performance of \OCKG with different kinds of change-points and graph structures. Two extra scenarios are included in the Appendix. 

The graph structures that we analyze are the Stochastic Block Models \citep{Holland1983} and the Barabási–Albert Model \citep{Reka2002}. In order to keep the simulation results comparable between random instances. We generate a fixed instance for each graph model and over it $50$ instances of each of the synthetic scenarios described bellow: 

\inlinetitle{\textup{I}.~Changes in node clusters}{.} 
We sample a Stochastic Block Model with $4$ clusters, $C1$,...,$C4$, each containing $20$ nodes. The intra- and inter-cluster node connection probability is fixed at $0.5$ and $0.01$, respectively.

\emph{Bivariate Gaussian distribution to Gaussian copula with uniform marginals}.  
In this first experiment, all nodes will follow a bivariate Gaussian model with the same covariance matrix and mean vector. Then we pick a cluster $C$ at time $t=2000$. From this moment, nodes of $C$ will generate observations from  a Gaussian copula ($\sim GC$) whose marginals follow uniform distributions ($\sim U(-c,c)$): 
\begin{equation}
\begin{aligned}
&(x,y) \sim N(\mu,\Sigma), \ \ \mu=(0,0), \ \ \Sigma_{x,x}=1,\Sigma_{x,y}=\frac{4}{5}  \ \ \rightarrow \\ 
&  (x,y) \sim GC, \ \ \Sigma_{x,x}=1,
\Sigma_{x,y}=\frac{4}{5}. \\
\end{aligned}
\end{equation}
The parameter $c$ is chosen so the mean vector and covariance matrix before and after the change-point are the same. This particular example is hard as the probabilistic model do not depend on the same set of parameters and the first two moments which are used for basic non-parametric methods are the same. 

\inlinetitle{\textup{II.}~Changes in a subset of connected nodes}{.} We generate a Barabási-Albert Model with $100$ nodes. We start with one node, and then one new node is added at each iteration $i$ and get connected with one of the nodes present at $i-1$ with a probability proportional to their degree. For each instance of the experiments, we will generate $C$ by selecting a node at random with probability proportional to its degree and all the nodes that are at a distance of $4$ in the graph. These nodes will suffer a change in the probability model generating its associated stream. 

\emph{Shift in the mean on one of the cluster components.} The streams observed at each of the components are drawn from a different multivariate Gaussian distribution of dimension $3$, before and after the change-point at time $\tau=1000$: 
%%%
\begin{equation}
\begin{aligned}
x_{v} &\sim N(\mu,\Sigma), \mu=\zeros{3}, \Sigma_{i,i}=1,\Sigma_{1,2}=\frac{4}{5},\Sigma_{3,1}=0 \ \ \rightarrow \\ 
x_{v} &\sim N(\mu,\Sigma), \mu=(1,0,0), \Sigma_{i,i}=1,\Sigma_{1,2}=\frac{4}{5},\Sigma_{3,1}=0.\!\!\!\!
\end{aligned}
\end{equation}
%%%
Further change-point detection scenarios are explored in the \Appendix{appendix:further experiment}

\setlength\tabcolsep{3 pt}
\begin{table*}[t]
%\begin{adjustbox}{width=15.0mm,center}
\footnotesize
\centering
\makebox[\linewidth][c]{%
\scalebox{.883}{
\begin{tabular}{c r || r r r ||| c r || r r r}
    \toprule
 \textbf{Scenario:} & & \textbf{Detection}  %($\downarrow$) 
&  %\textbf{AUC}  %($\downarrow$) 
&     &  \textbf{Scenario:}  %($\downarrow$) 
&  %\textbf{AUC}  %($\downarrow$) 
&  \textbf{Detection}
\\
\textbf{Experiment I\!} & \textbf{Detector} & \textbf{delay (std)}  %($\downarrow$) 
&   \textbf{AUC (std)}  %($\downarrow$) 
&  \textbf{Precision}  & \textbf{Experiment II\!} & \textbf{\ Detector} & \textbf{delay (std)}  %($\downarrow$) 
&   \textbf{AUC (std)}  %($\downarrow$) 
&  \textbf{Precision} 
\\
    \hline\hline
%\emph{Synthetic}:   
& \OCKG $\alpha=$ 0.1 & 126.26 \ \   (11.95)  & \textbf{0.89 (0.05)}  &  \textbf{1.00} \ \ & 
%\ \  \  \emph{Synthetic }:   
& \OCKG $\alpha=$ 0.1 & 25.44  \ \ (1.96) & \textbf{0.97 (0.02)} & \textbf{1.00}  \ \  \\
%\textbf{Experiment I} 
&  \ \OCKG $\alpha=$ 0.5 & 129.67  \ \  (11.37) &  0.85 (0.06) &  0.98 \ \ & 
%Experiment II 
&     \ \OCKG $\alpha=$ 0.5 & 25.06 \ \ (1.34) & \textbf{0.97 (0.02)}  & 0.96 \ \ \\
$n=$125 &      \  Pool $\alpha=$ 0.1   &  123.72  \ \   (24.08) & 0.82 (0.05)  & 0.58 \ \ &  $n=$25 &      \  Pool $\alpha=$ 0.1   &  24.51  \ \ (1.68) & 0.91 (0.03)  & 0.82 \ \ \\
    &      \  Pool $\alpha=$ 0.5   & 131.90  \ \   (21.02)  & 0.80 (0.05)  & 0.80 \ \ &  &  \  Pool $\alpha=$ 0.5   & 24.44  \ \ (2.03)  & 0.93 (0.02) &  0.86 \ \  \\
       &   Nougat   & 146.50  \ \  (70.74)  & 0.56 (0.22)  &  0.12  \ \  &  &   Nougat   & 34.25 (13.80)  & 0.64 (0.17)   &   0.08  \ \ \\ 
\hline 
%\emph{Synthetic}:   
& \OCKG $\alpha=$ 0.1 & 252.72 \ \ (14.82) & \textbf{0.93 (0.03)}  & \textbf{1.00} \ \ & 
%\emph{Synthetic}:   
& \OCKG $\alpha=$ 0.1 & 50.38  \ \ (1.21)  & \textbf{0.99 (0.01)}  &     \textbf{1.00}  \ \  \\  
%\textbf{Experiment I}  
&     \ \OCKG $\alpha=$ 0.5 & 251.31 \ \ (21.82) & 0.89 (0.04) & 0.98 \  \ & 
%\textbf{Experiment II} 
&     \ \OCKG $\alpha=$ 0.5 & 50.67  \ \ (1.48) &  0.96 (0.04) &  0.98 \  \  \\
$n=$250 &      \  Pool $\alpha=$ 0.1   &  249.54 \ \ (25.20)  & 0.86 (0.04)  & 0.92 \ \ &  $n=$50 &      \  Pool $\alpha=$ 0.1   & 48.55  \ \ (5.14)  &  0.91 (0.04) & 0.98  \  \  \\
    &      \  Pool $\alpha=$ 0.5   & 245.32  \ \ (22.20) & 0.87 (0.04) & 0.94  \  \  &    &  Pool $\alpha=$ 0.5   &  49.63  \ \ (2.38) & \textbf{0.99 (0.01)}  &   0.98 \  \ \\  
  & Nougat   &  273.50 \ \ (96.59) & 0.67 (0.19)  & 0.20   \  \  &   & Nougat   & 77.84 (10.24) & 0.75 (0.17)  &  0.76 \  \ \\
\hline
% \emph{Synthetic}:   
& \ \OCKG $\alpha=$ 0.1 & 502.70 \ \ \ \ (6.49)  & \textbf{0.99 (0.00)}   & \textbf{1.00} \  \  &  
%\emph{Synthetic}:   
& \OCKG $\alpha=$ 0.1 &  100.52  \ \ (1.25) & 0.99 (0.00)  &  \textbf{1.00} \ \  \\
%\textbf{Experiment II}  
&    \ \OCKG $\alpha=$ 0.5 & 500.84 \ \ \ \ (5.15) & \textbf{0.99 (0.00)}  & \textbf{1.00}  \ \  & 
%Experiment II 
&     \ \OCKG $\alpha=$ 0.5 & 100.16  \ \ (0.64)  & \textbf{1.00 (0.00)}  & \textbf{1.00} \ \    \\
$n=$500 &      \  Pool $\alpha=$ 0.1   & 506.20 \ \ (18.31)  & \textbf{0.99 (0.00)} & \textbf{1.00}  \  \ & $n=$100 &      \  Pool $\alpha=$ 0.1   & 99.86  \ \ (1.23)  &  0.99 (0.00) & \textbf{1.00}  \ \  \\
    &      \  Pool $\alpha=$ 0.5   & 501.90 \ \ \ \ (7.86)   &  \textbf{0.99 (0.00)}  &  \textbf{1.00}  \  \ &  & \  Pool $\alpha=$ 0.5   & 100.38  \ \ (1.01)  &  0.99   (0.00) & \textbf{1.00} \ \  \\
       &  \  Nougat   &  576.86 (129.27) &  0.66 (0.20) & 0.74 \  \ &  &  \  Nougat   & 127.52  \ (13.87)  & 0.77 (0.16)   &  0.88 \ \   \\
\hline
 \bottomrule
\end{tabular}
}
}
\vspace{-2mm}
\caption{Performance comparison between change-point detectors in three synthetic scenarios for different window sizes ($n$ value). The mean and standard deviation of the score is based on $50$ instances.}\label{tab:synthetic experiments}
%\end{adjustbox}
\end{table*}

\begin{figure*}[t]
\centering
%\ \ \ \ \includegraphics[width=0.045\textwidth, viewport=435 235 575 280, clip]{./random.pdf}\\
%\vspace{-3mm}
\!\!\!\!\!\!\!\!\!\subfigure{
}
\vspace{-2mm}
\centering
\subfigure{\includegraphics[width=0.245\textwidth]{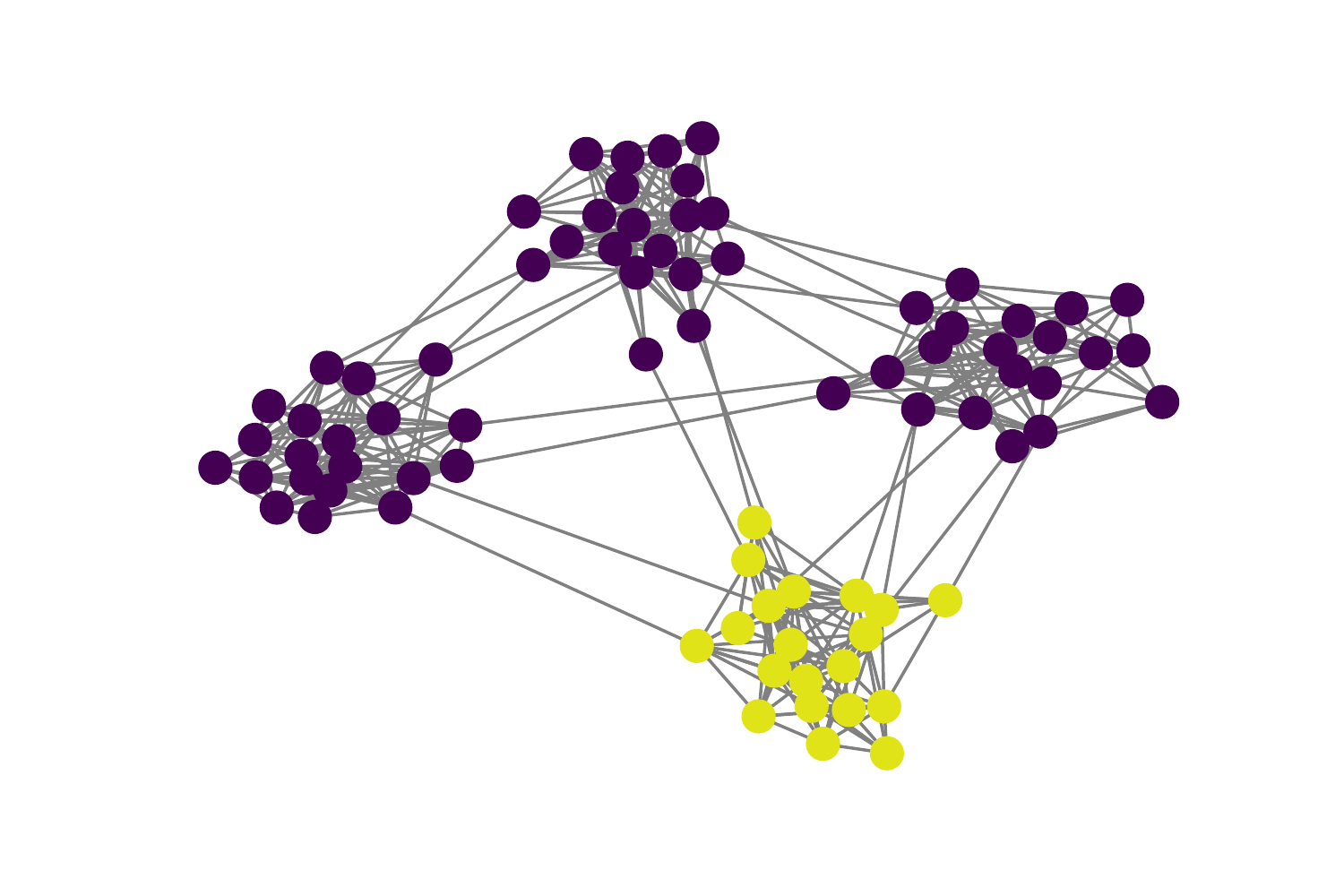}
\includegraphics[width=0.245\textwidth]{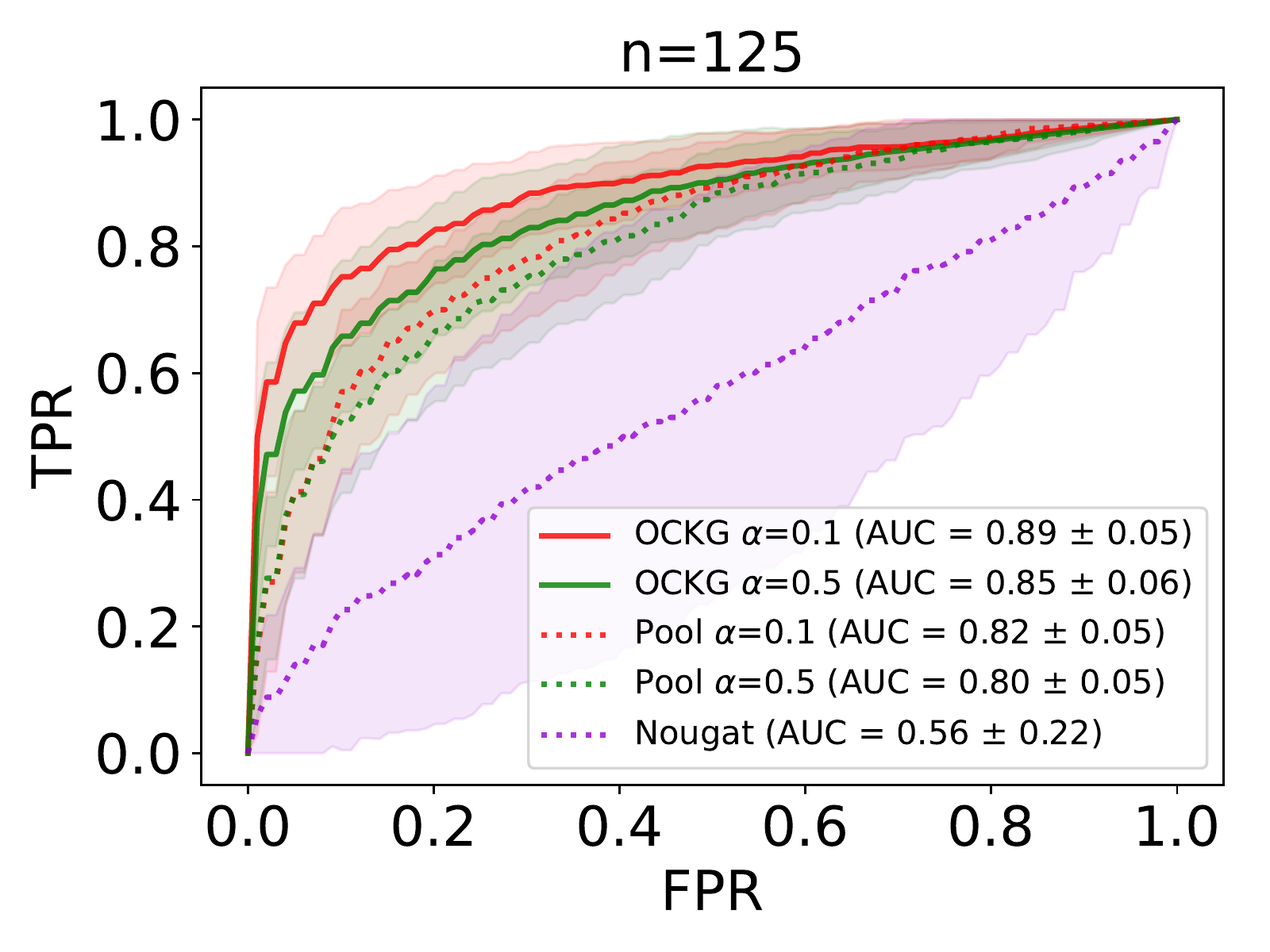}
\includegraphics[width=0.245\textwidth]{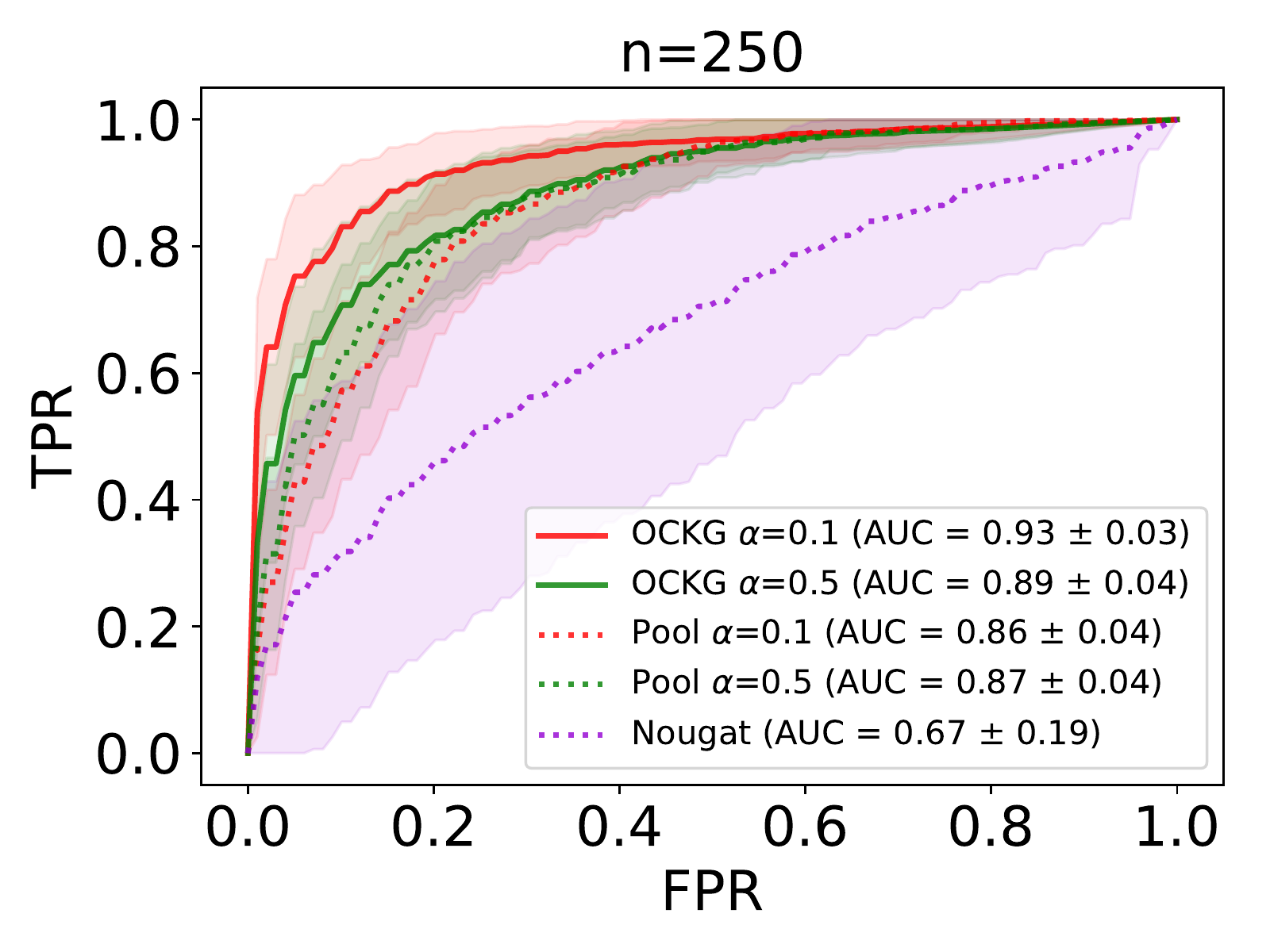}
\includegraphics[width=0.245\textwidth]{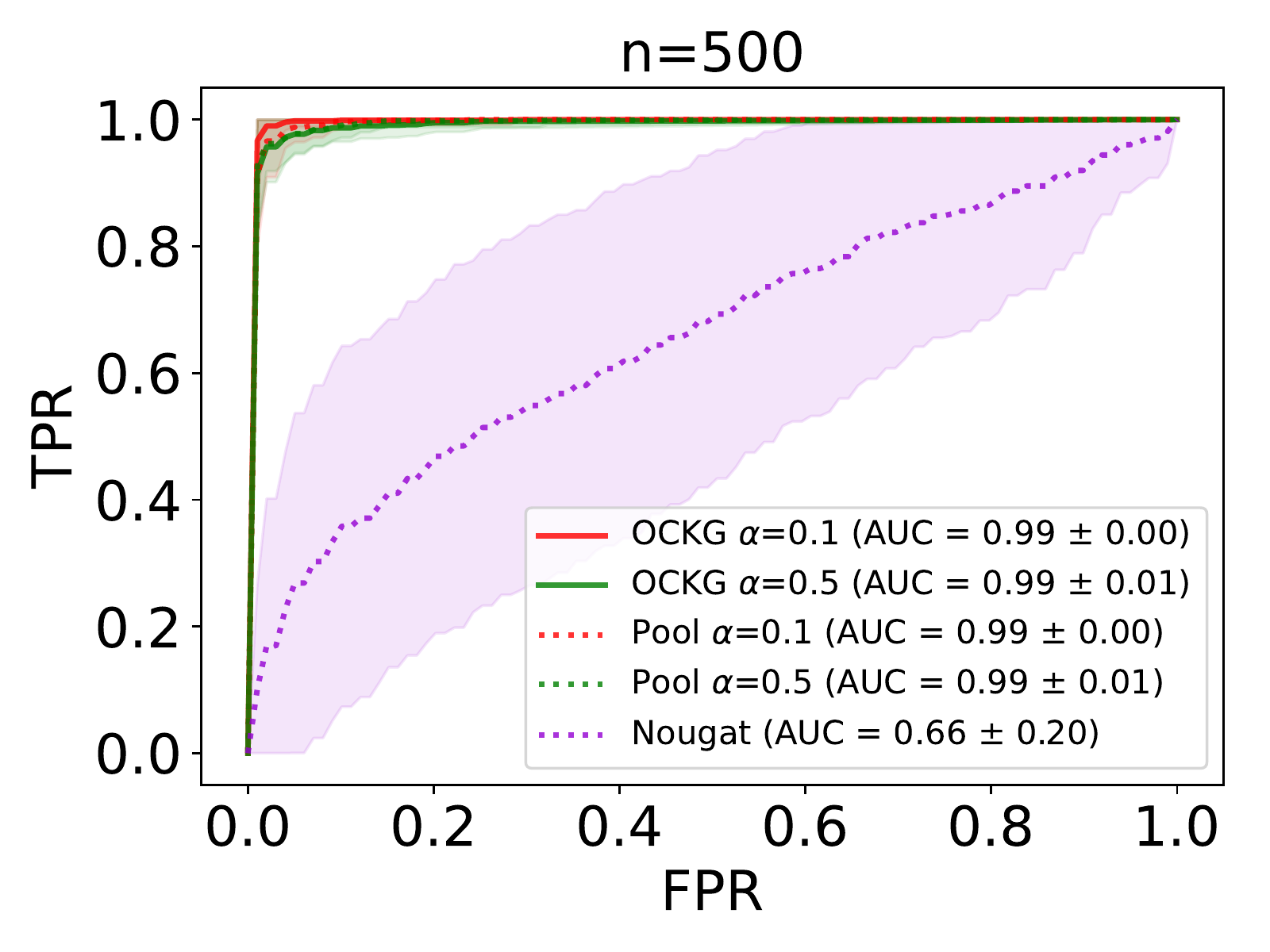}
}
\vspace{-2mm}
\centering
\subfigure{
\includegraphics[width=0.245\textwidth]{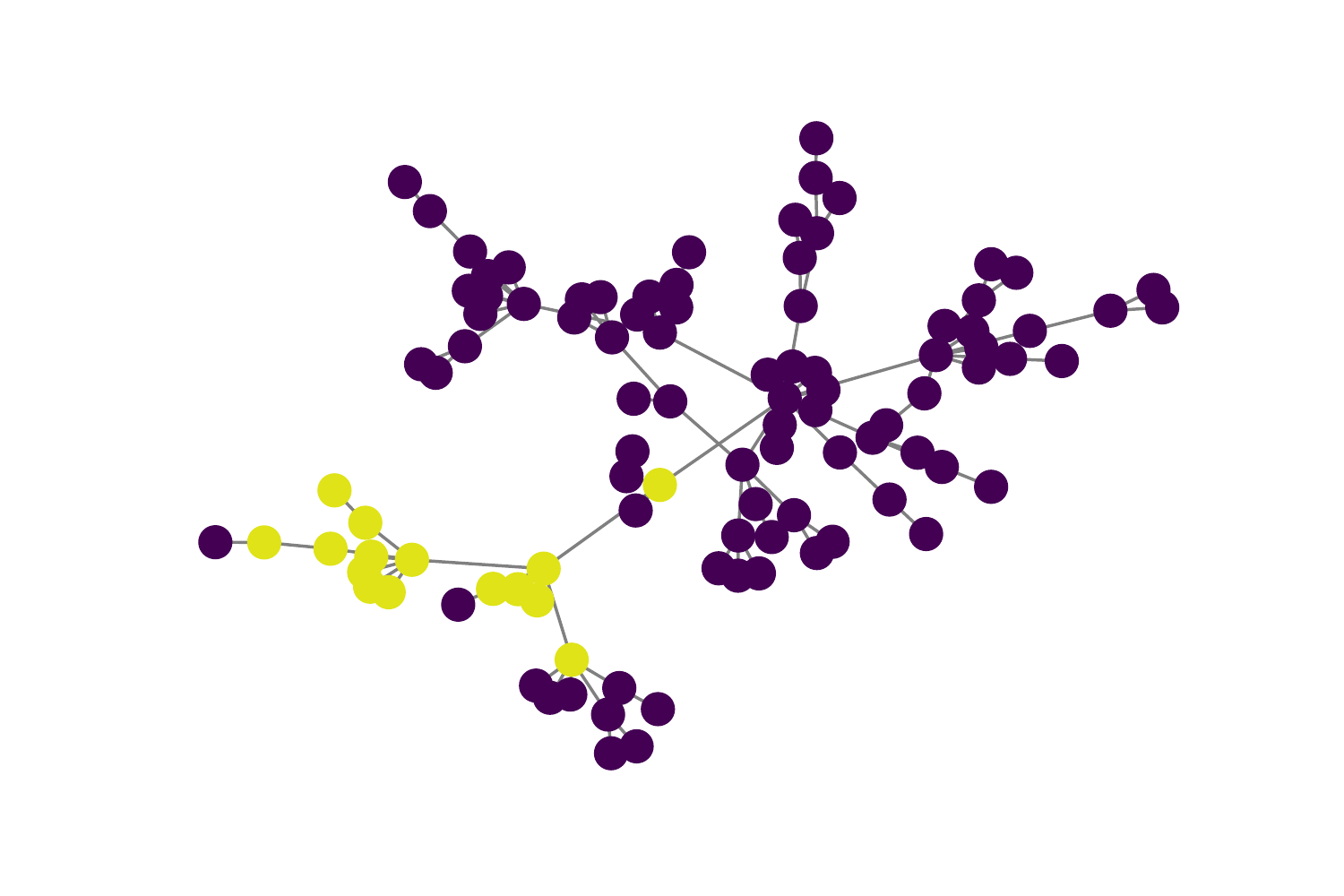}
\includegraphics[width=0.245\textwidth]{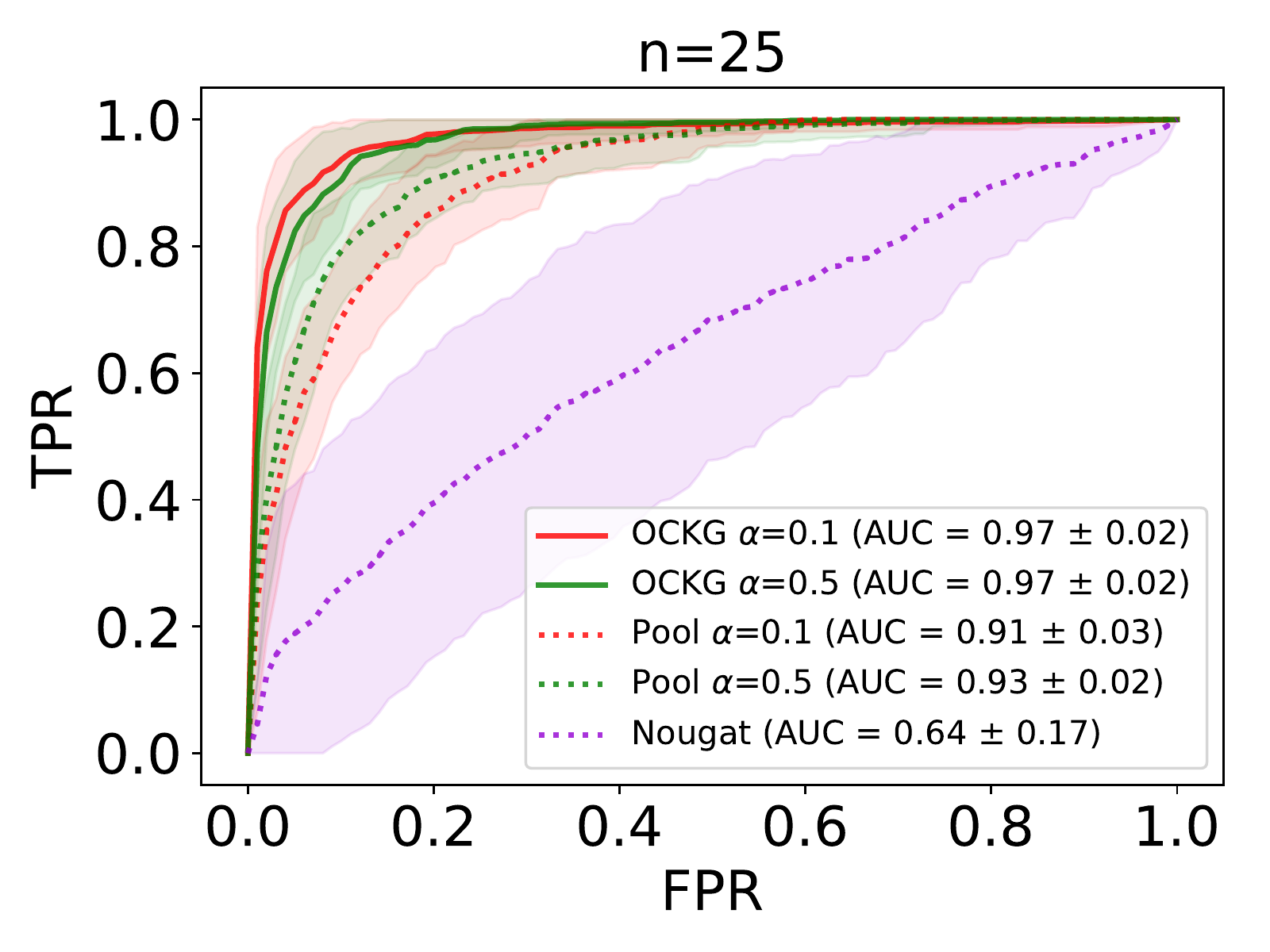}
\includegraphics[width=0.245\textwidth]{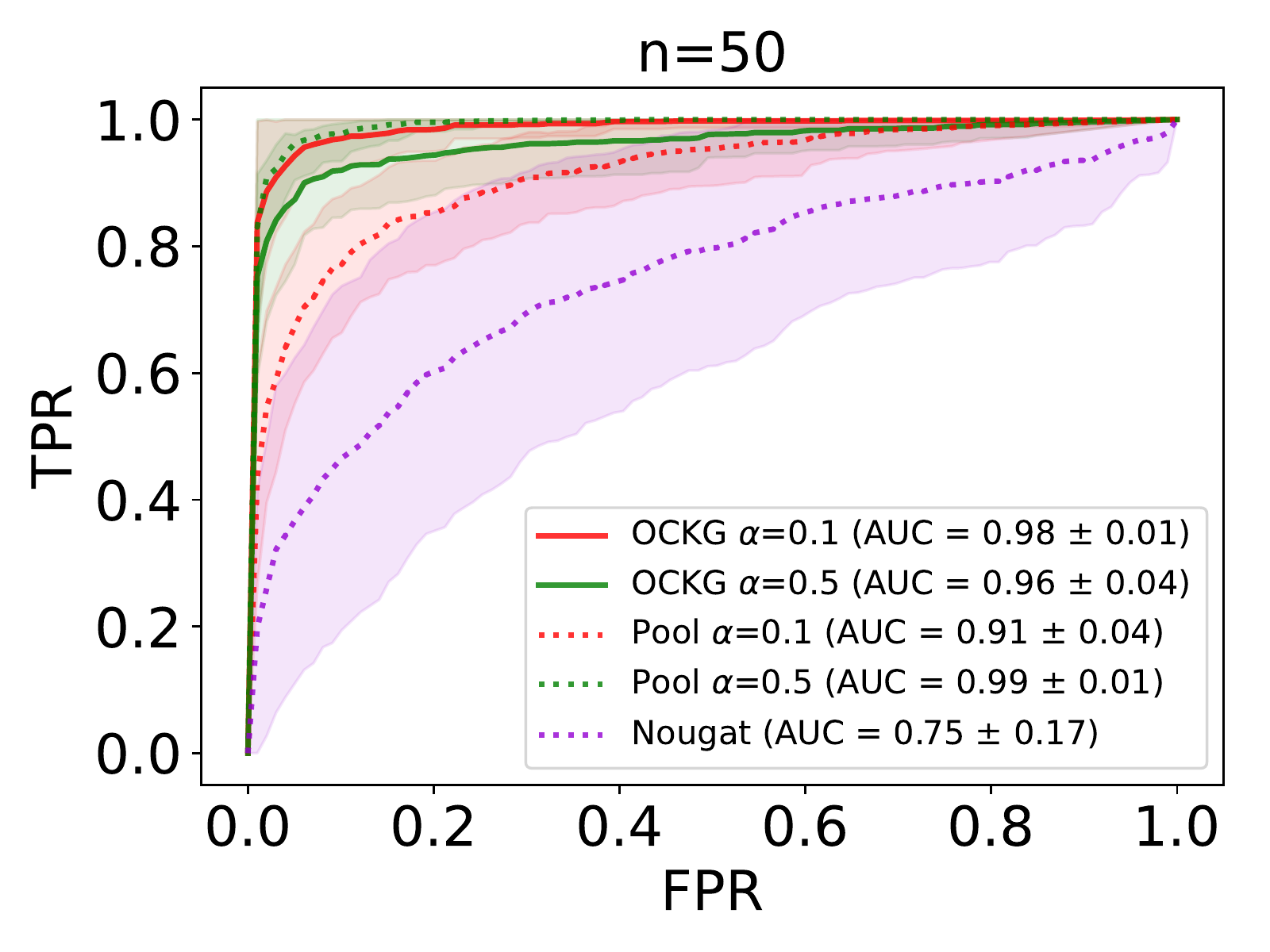}
\includegraphics[width=0.245\textwidth]{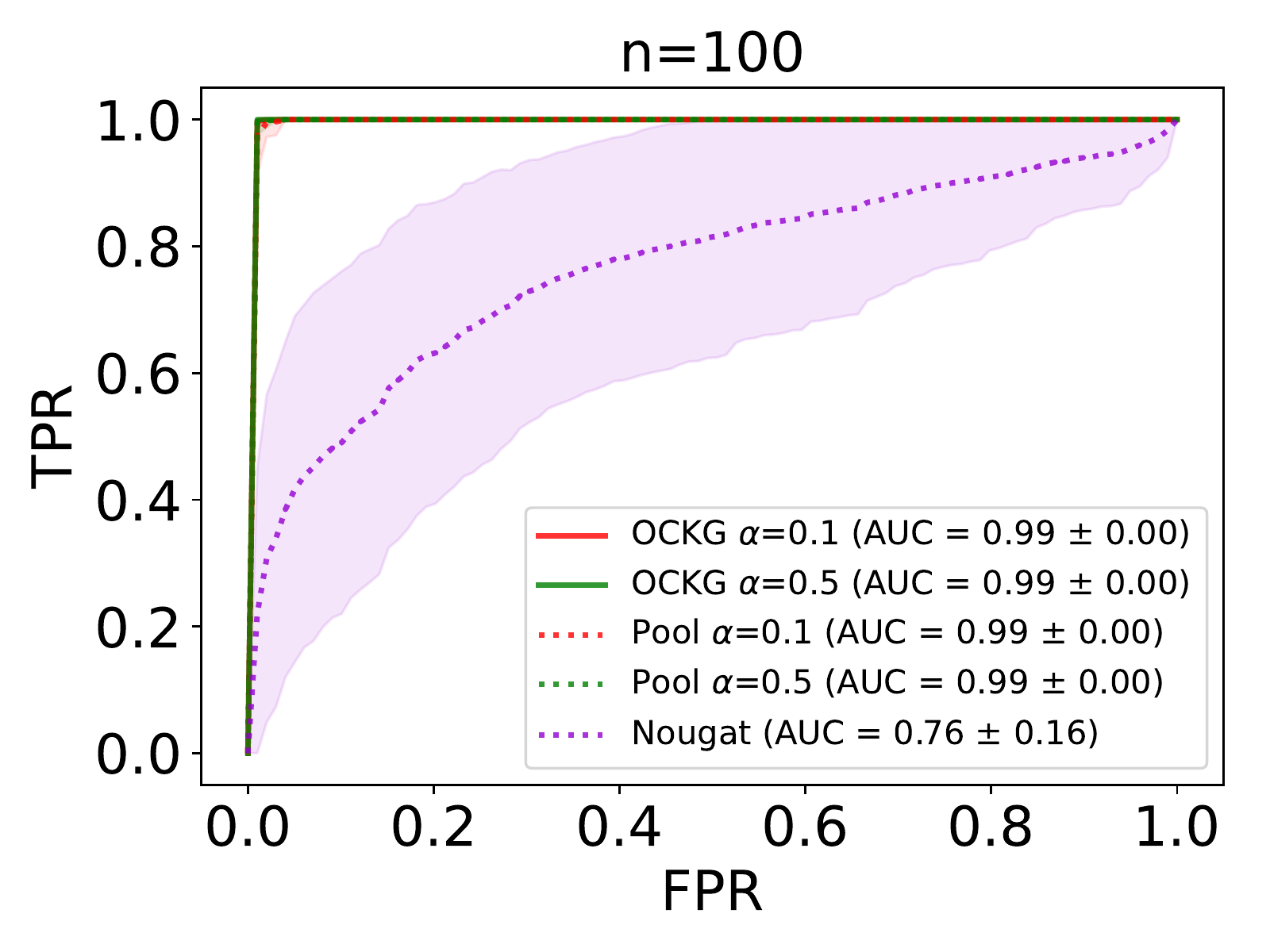}}
\vspace{-2mm}
\vspace{-2mm}
\caption{Each row presents results for the two synthetic scenarios. In the first column, an instance of the simulated nodes that suffer a change-point is shown (yellow nodes). The next columns show the mean ROC curves along with their standard deviations for different window sizes $n$. The mean and standard deviations are estimated based on $50$ random instances.
-- \emph{Experiment I}: A bivariate Gaussian distribution changes to Gaussian copula with uniform marginals. The change affects one cluster of nodes. -- 
\emph{Experiment II}: Change in the mean in one dimension of a 3d Gaussian distribution. The change affects a small set of connected nodes. 
}\label{fig:results}
\end{figure*}

\inlinetitle{Discussion on the results}{.}~%
The results of the synthetic experiments are summarized in \Tab{tab:synthetic experiments} and \Fig{fig:results}. As it was explained before, the method derived from the non-parametric LRE lack a solid theoretical and computationally efficient methodology to fix the parameter related with the detection of the change point, $\eta$ and $\eta_v$. Nevertheless, the related scores are expected to achieve its highest value around $\tau+n$. In order to make fair comparison between methods, we report the expected delay based on the peak observed in the time-series of the global score generated by each method. Similarly, we compare the methods in terms of the averaged ROC curves of the classification problem aiming to detect the set of nodes that are chosen in each synthetic scenario, such curves are based on the node level scores observed at time $\tau+n$. In \Tab{tab:synthetic experiments} we report the averaged AUC of the ROC curves and its standard deviation, as well as the percentage of times when the changepoint $\tau$ was successfully identified.

We can see in \Tab{tab:synthetic experiments} and \Fig{fig:results} that the performance of the algorithms improve as the number of observations increases in general, which is not surprising. We can see that the Nougat method requires the largest amount of observations to identify $\tau$ and the set $C$. We believe that is due to the stochastic gradient descent step, which produces more noisy detection scores when compared with other methods. In most synthetic scenarios, \OCKG-based variants have the best performance in situations where fewer observations are available, as it can be seen in the related precision and AUC scores. Then, by exploiting the graph component, a method enhances the precision of the change-point detection and lowers the detection delay. With respect to the parameter $\alpha$, the experiments suggest the use of smaller values, such as $\alpha=0.1$, to keep the detector sensitive to situations in which the Pearson's Divergence between the pre- and post-change \pdfuncs is expected to be small.%%%
\section{Conclusions}
%%%
In this paper, we introduced a framework to detect change-points in multivariate streams over the nodes of a graph. The main appeal of our approach is its non-parametric formulation that integrates the a priori provided information of the graph structure, and is able to spot different types of change-point with minimum hypotheses. 

In the experiments, we provided evidence showing how the Laplacian regularization %term 
leads to an online detector with a lower expected detection time. Moreover, this compares nicely against other alternatives, such as the Nougat method, which also exploits the graph structure. Moreover, the formulation of the cost function based on the Pearson's Divergence leads to an elegant optimization problem with convenient convergence guarantees. 

There are still some questions that remain to be answered and can be part of future work: the selection of the threshold parameters, which is an open problem in general for non-parametric likelihood-ratio based methods, and requires a broader theoretical analysis; the time-dependency of the data; and cases where there is a delay in how each node reacts to a change-point.

\section{Acknowledgments}\label{sec:acknowledgments}

This work was supported by the Industrial Data Analytics and Machine Learning (IdAML) Chair hosted at ENS Paris-Saclay, University Paris-Saclay, and grants from Région Ile-de-France.

\bibliography{References} 

\begin{thebibliography}{}

\bibitem[Albert and Barab\'asi, 2002]{Reka2002}
Albert, R. and Barab\'asi, A.-L. (2002).
\newblock Statistical mechanics of complex networks.
\newblock {\em Rev. Mod. Phys.}, 74:47--97.

\bibitem[Aminikhanghahi and Cook, 2017]{Aminikhanghahi2017}
Aminikhanghahi, S. and Cook, D.~J. (2017).
\newblock Using change point detection to automate daily activity segmentation.
\newblock In {\em IEEE \International \Conference on Pervasive Computing and
  Communications Workshops}, pages 262--267.

\bibitem[Arlot et~al., 2019]{Arlot2019}
Arlot, S., Celisse, A., and Harchaoui, Z. (2019).
\newblock A kernel multiple change-point algorithm via model selection.
\newblock {\em \Journal of Machine Learning Research}, 20(162):1--56.

\bibitem[Beck and Tetruashvili, 2013]{Beck2013}
Beck, A. and Tetruashvili, L. (2013).
\newblock On the convergence of block coordinate descent type methods.
\newblock {\em SIAM \Journal on Optimization}, 23:2037--2060.

\bibitem[Bouchikhi et~al., 2019]{Bouchikhi2019}
Bouchikhi, I., Ferrari, A., Richard, C., Bourrier, A., and Bernot, M. (2019).
\newblock Kernel based online change point detection.
\newblock In {\em European Signal Processing \Conference}, pages 1--5.

\bibitem[Csiszár, 1967]{Csiszar1967}
Csiszár, I. (1967).
\newblock On topological properties of f-divergences.
\newblock {\em Studia Scientiarum Mathematicarum Hungarica}, 2:329--339.

\bibitem[de~la Concha et~al., 2022]{delaConcha2022}
de~la Concha, A., Kalogeratos, A., and Vayatis, N. (2022).
\newblock Collaborative likelihood-ratio estimation over graphs.

\bibitem[{Ferrari} and {Richard}, 2020]{Ferrari2020}
{Ferrari}, A. and {Richard}, C. (2020).
\newblock Non-parametric community change-points detection in streaming graph
  signals.
\newblock In {\em IEEE \International \Conference on Acoustics, Speech and
  Signal Processing}, pages 5545--5549.

\bibitem[Ferrari et~al., 2021]{Ferrari2021}
Ferrari, A., Richard, C., Bourrier, A., and Bouchikhi, I. (2021).
\newblock Online change-point detection with kernels.

\bibitem[Harchaoui et~al., 2008]{Harchaoui2008}
Harchaoui, Z., Moulines, E., and Bach, F. (2008).
\newblock Kernel change-point analysis.
\newblock In Koller, D., Schuurmans, D., Bengio, Y., and Bottou, L., editors,
  {\em Advances in Neural Information Processing Systems}, volume~21. Curran
  Associates, Inc.

\bibitem[Holland et~al., 1983]{Holland1983}
Holland, P.~W., Laskey, K.~B., and Leinhardt, S. (1983).
\newblock Stochastic blockmodels: First steps.
\newblock {\em Social Networks}, 5(2):109--137.

\bibitem[Kanamori et~al., 2012]{Kanamori2012}
Kanamori, T., Suzuki, T., and Sugiyama, M. (2012).
\newblock f-divergence estimation and two-sample homogeneity test under
  semiparametric density-ratio models.
\newblock {\em IEEE \Transactions on Information Theory}, 58(2):708--720.

\bibitem[Kawahara and Sugiyama, 2012]{Kawahara2012}
Kawahara, Y. and Sugiyama, M. (2012).
\newblock Sequential change-point detection based on direct density-ratio
  estimation.
\newblock {\em Statistical Analysis and Data Mining: The ASA Data Science
  \Journal}, 5(2):114--127.

\bibitem[Kullback, 1959]{Kullback59}
Kullback, S. (1959).
\newblock {\em Information Theory and Statistics}.
\newblock Wiley.

\bibitem[Li et~al., 2019]{Li2019}
Li, S., Xie, Y., Dai, H., and Song, L. (2019).
\newblock Scan {B}-statistic for kernel change-point detection.
\newblock {\em Sequential Analysis}, 38(4):503--544.

\bibitem[Li et~al., 2018]{Li2018}
Li, X., Zhao, T., Arora, R., Liu, H., and Hong, M. (2018).
\newblock On faster convergence of cyclic block coordinate descent-type methods
  for strongly convex minimization.
\newblock {\em \Journal of Machine Learning Research}, 18(184):1--24.

\bibitem[Nguyen et~al., 2008]{Nguyen2007}
Nguyen, X., Wainwright, M.~J., and Jordan, M. (2008).
\newblock Estimating divergence functionals and the likelihood ratio by
  penalized convex risk minimization.
\newblock In {\em Advances in Neural Information Processing Systems}.

\bibitem[Page, 1954]{Page1954}
Page, E.~S. (1954).
\newblock {Continuous inspection schemes}.
\newblock {\em Biometrika}, 41(1--2):100--115.

\bibitem[Pearson, 1900]{Pearson1900}
Pearson, K. (1900).
\newblock X. {O}n the criterion that a given system of deviations from the
  probable in the case of a correlated system of variables is such that it can
  be reasonably supposed to have arisen from random sampling.
\newblock {\em The London, Edinburgh, and Dublin Philosophical Magazine and
  Journal of Science}, 50(302):157--175.

\bibitem[Richard et~al., 2009]{Cedric2009}
Richard, C., Bermudez, J. C.~M., and Honeine, P. (2009).
\newblock Online prediction of time series data with kernels.
\newblock {\em IEEE \Transactions on Signal Processing}, 57(3):1058--1067.

\bibitem[{Sharpnack} et~al., 2016]{Sharpnack2016}
{Sharpnack}, J., {Rinaldo}, A., and {Singh}, A. (2016).
\newblock Detecting anomalous activity on networks with the graph fourier scan
  statistic.
\newblock {\em IEEE \Transactions on Signal Processing}, 64(2):364--379.

\bibitem[Sheldon, 2008]{Sheldon2008}
Sheldon, D. (2008).
\newblock {Graphical Multi-Task Learning}.
\newblock Technical report, Cornell University.

\bibitem[Shewhart, 1925]{Shewart1925}
Shewhart, W.~A. (1925).
\newblock The application of statistics as an aid in maintaining quality of a
  manufactured product.
\newblock {\em \Journal of the American Statistical Association},
  20(152):546--548.

\bibitem[Shiryaev, 1963]{Shiryaev1963}
Shiryaev, A.~N. (1963).
\newblock On optimum methods in quickest detection problems.
\newblock {\em Theory of Probability \& Its Applications}, 8(1):22--46.

\bibitem[Shuman et~al., 2013]{Shuman2013}
Shuman, D.~I., Narang, S.~K., Frossard, P., Ortega, A., and Vandergheynst, P.
  (2013).
\newblock The emerging field of signal processing on graphs: {E}xtending
  high-dimensional data analysis to networks and other irregular domains.
\newblock {\em {IEEE} Signal Processing Magazine}, 30(3):83--98.

\bibitem[Sugiyama et~al., 2007]{Sugiyama2007}
Sugiyama, M., Nakajima, S., Kashima, H., Buenau, P., and Kawanabe, M. (2007).
\newblock Direct importance estimation with model selection and its application
  to covariate shift adaptation.
\newblock In {\em Advances in Neural Information Processing Systems}.

\bibitem[Sugiyama et~al., 2011]{Sugiyama2011}
Sugiyama, M., Suzuki, T., Itoh, Y., Kanamori, T., and Kimura, M. (2011).
\newblock Least-squares two-sample test.
\newblock {\em Neural networks}, 24:735--51.

\bibitem[Tartakovsky, 2021]{Tartakovsky2021}
Tartakovsky, A. (2021).
\newblock {\em Sequential change detection and hypothesis testing : general
  non-i.i.d. stochastic models and asymptotically optimal rules}.
\newblock Chapman \& Hall/CRC.

\bibitem[Tartakovsky et~al., 2014]{Tartakovsky2014}
Tartakovsky, A., Nikiforov, I., and Basseville, M. (2014).
\newblock {\em Sequential Analysis: Hypothesis Testing and Changepoint
  Detection}.
\newblock Chapman \& Hall/CRC Monographs on Statistics \& Applied Probability.
  Taylor \& Francis, CRC Press.

\bibitem[Xie et~al., 2021]{Xie2021}
Xie, L., Zou, S., Xie, Y., and Veeravalli, V.~V. (2021).
\newblock Sequential (quickest) change detection: Classical results and new
  directions.
\newblock {\em IEEE \Journal on Selected Areas in Information Theory}.

\bibitem[Yamada et~al., 2011]{Yamada2011}
Yamada, M., Suzuki, T., Kanamori, T., Hachiya, H., and Sugiyama, M. (2011).
\newblock Relative density-ratio estimation for robust distribution comparison.
\newblock In {\em Advances in Neural Information Processing Systems}.

\bibitem[Zou and Veeravalli, 2018]{Zou2018}
Zou, S. and Veeravalli, V.~V. (2018).
\newblock Quickest detection of dynamic events in sensor networks.
\newblock In {\em IEEE \International \Conference on Acoustics, Speech and
  Signal Processing}, pages 6907--6911.

\bibitem[Zou et~al., 2019]{Zou2019}
Zou, S., Veeravalli, V.~V., Li, J., Towsley, D., and Swami, A. (2019).
\newblock Distributed quickest detection of significant events in networks.
\newblock In {\em IEEE \International \Conference on Acoustics, Speech and
  Signal Processing}, pages 8454--8458.

\end{thebibliography}

\newpage

\newpage

\onecolumn

\appendix

\section*{APPENDIX}

\section{Analysis of the proposed optimization algorithm}\label{appendix:optimization}

In this section, we discuss the details of the Cyclic Block Gradient Descent strategy described in \Sec{sec:practical_implementation}. In particular, we are interested in proving an upper bound for the number of interactions to attain a given precision $\epsilon$ in terms of the size of the dictionary and the number of nodes. We use $\eigmax{A}$ to indicate the maximum eigenvalue of the square matrix A.

\begin{theorem}\label{Th:convergence}
Suppose that for a dictionary $D$ of size $L \geq 2$ we aim to solve the Optimization \Problem{eq:thetas_cost_function} via the Cyclic Block Gradient Descent strategy, where the update with respect to the node parameter $\theta_v$ at the $i$-th cycle is computed by:
\begin{equation}\label{eq:update_2}
\begin{aligned} 
& \hthetanode{v,t}^{(i)}   \mydef  \frac{\eta_{v,t}}{\eta_{v,t} + \lambda \gamma} \hthetanode{v,t}^{(i-1)} - \frac{1}{\eta_{v,t} + \lambda \gamma}
 \Bigg[ \left(  \frac{(1-\alpha) H_{v,t} + \alpha H'_{v,t}}{\Gdims} \right) \hthetanode{v,t}^{(i-1)} - \frac{h'_{v,t}}{\Gdims} \Bigg] \\ 
& - \frac{\lambda}{\eta_{v,t} + \lambda \gamma} \left[ d_v \hthetanode{v,t}^{(i-1)} - \sum_{u \in \nghood(v)\!\!\!\!\!\!\!\!} W_{uv} \big( \hthetanode{u,t}^{(i)} \one_{u<v} + \hthetanode{u,t}^{(i-1)} \one_{u \geq v} \big) \right].
\end{aligned}
\end{equation}
Then, if we fix the learning rate for node $v$ at $\eta_{v,t}= \eigmax{\!\frac{ (1-\alpha)}{\Gdims} H_{v,t} + \frac{\alpha}{\Gdims} H'_{v,t} + \lambda d_v \id_L \!}$, we will need at most the following number of interactions in order to achieve a pre-specified accuracy level $\epsilon>0$:
\begin{equation}{\label{eq:complexity}}
\begin{aligned}
& i_{\textup{max}}  \mydef  \Bigg\lceil \frac{\lambda \gamma (M_{\textup{min}}+\lambda \gamma) + 16 M^2 \log^2(3 \Gdims L)}{\lambda \gamma (M_{\textup{min}}+\lambda \gamma)}\log\bigg( \frac{\Phi(\Theta^{(0)}_{t})-\Phi(\Theta^*_{t})}{\epsilon}\bigg) \Bigg\rceil, 
\end{aligned}
\end{equation} 
where: 
\begin{equation}
    \Phi(\Theta)= \frac{1}{\Gdims} \sum_{v \in V} \ell_{v,t}(\theta_v) +  \frac{\lambda}{2} \Theta^\top ([\Lpc+\gamma \id_{\Gdims}] \otimes \id_L) \Theta
\end{equation}
\begin{equation}
\begin{aligned}
& M \mydef \eigmax{ \frac{1-\alpha}{\Gdims} H_{t}  + \frac{\alpha}{\Gdims} H'_{t} +  \lambda [\Lpc \otimes \id_L]}\!, \ \ M_{v} \mydef   \eigmax{ \frac{(1-\alpha)}{\Gdims}  H_{v,t}+ \frac{\alpha}{\Gdims} H'_{v,t} + \lambda d_v \id_L }\!, \ \ M_{\textup{min}}  \mydef \min_{v \in V} M_v. 
\end{aligned}
\end{equation}
\end{theorem}

\Theorem{Th:convergence} is a particular case of the results appearing in the paper \cite{Li2018}. In that work, the authors analyzed the convergence Cyclic Block Coordinate type-algorithms. For completeness of the presentation, we present some of their main results. 

The objective functions analyzed in \cite{Li2018} take the form: 
\begin{equation}{\label{eq:LU_opt_problem}}
\begin{aligned}
    \min_{\Theta \in \R^M} \Phi(\Theta) &=  \min_{\theta \in \R^M}  Q(\Theta) + R(\Theta),
\end{aligned}
\end{equation}
where $Q$ is a twice differentiable loss function, $R$ is a possibly non-smooth and strongly convex penalty function and the variable $\Theta$ is of dimension $L=\sum_{v=1}^{\Gdims} L_v$ and is partitioned into disjoint blocks, $\Theta=(\theta_1,\theta_2,...,\theta_\Gdims)$, each of them being of dimension $M_v$. It is supposed the penalization term can be written as $R(\Theta)=\sum_{v=1}^\Gdims R_v(\theta_v)$.

\begin{assumption_2}\label{ass_a:Lipschitz}
$Q(\cdot)$ is convex, and its gradient mapping $\Delta Q (\cdot)$ is Lipschitz-continuous and also block-wise Lipschitz-continuous, \ie there exist positive constants $M$ and $M_v$ such that for any $\Theta,\Theta' \in \R^M$ and $v=1,...,\Gdims$, we have:%\mN{$C$ is the cluster in the experiments}
\begin{equation}{\label{eq:Li_problem}}
\begin{aligned}
\norm{\Delta Q(\Theta') - \Delta Q(\Theta)}  & \leq M \norm{\Theta'-\Theta} \\
\norm{\Delta_v Q(\theta'_{u<v},\theta_v,\theta'_{u>v}) - \Delta_v Q(\Theta')} &\leq M_v \norm{\theta_v-\theta'_v}.
\end{aligned}
\end{equation}
\end{assumption_2}
%
%where $\Xi_{k-1}$ refers to the set of variables sampled at the interaction $k-1$

\begin{assumption_2}\label{ass_a:Lipschitz_r}
$R(\cdot)$ is strongly convex and also block-wise strongly convex, i.e. there exist positive constants $\mu$ and $\mu_v's$ such that for any $\Theta,\Theta' \in \R^M$ and  $v\in V$, we have:
\begin{equation}
\begin{aligned}
R(\Theta) & \geq R(\Theta') +(\Theta-\Theta')^{\top}\xi + \frac{\mu}{2}\norm{\Theta-\Theta'}^2, \\
R_v(\theta_v) & \geq R(\theta'_v) +(\theta_v-\theta'_v)^{\top}\xi_v + \frac{\mu_v}{2}\norm{\theta_v-\theta'_v}^2,
\end{aligned}
\end{equation}
for all $\xi \in \Delta R(\Theta')$.
\end{assumption_2}
Under the aforementioned hypothesis, the Cyclic Block Gradient Descent method, in which the cycle $i$ for block $v$, is defined as: 
\begin{equation}\label{eq:CCBGD_update}
\begin{aligned} 
\hthetanode{v}^{(i)} &  \mydef  \argmin_{\theta_v} (\theta_v - \hthetanode{v}^{(i-1)})^T \Delta_v Q (\hthetanode{u<v}^{(i)}, \hthetanode{u \geq v}^{(i-1)}) + \frac{\eta_v}{2} \norm{\theta_v-\hthetanode{v}^{(i-1)}}^2 + R_v(\theta_v).
\end{aligned}
\end{equation}
Then, Theorem\,A.1 characterizes the maximum number of interaction required to achieve a pre-specified accuracy $\epsilon$.

\begin{theorem_2}\label{Th:Li2018}
(Theorem 3 in \cite{Li2018}) -- Suppose that Assumptions 1 and 2 hold with $L \geq 2$. And that the optimization point is $\Theta^*$. We choose $\alpha_v=M_v$ for the CBC method. Given a pre-specified accuracy $\epsilon>0$ of the objective value, we need at most % 
\begin{equation}
i_{\textup{max}}  \mydef  \bigg\lceil \frac{\mu M_{\textup{min}}^{\mu} + 16 M^2 \log^2(3 \Gdims L_{\textup{max}})}{\mu M_{\textup{min}}^{\mu}} \log( \frac{\phi(\Theta^{(0)})-\phi(\Theta^*)}{\epsilon}) \bigg\rceil 
\end{equation}
iterations to ensure $\phi(\Theta^{(i)})-\phi(\Theta^*) < \epsilon $ for $i \geq i_{\textup{max}}$, where $M_{\textup{min}}^{\mu} \mydef \min_{v \in V} M_v + \mu_v$ and $L_{\textup{max}}=\max_{v \in V } L_v$.
\end{theorem_2}
\inlinetitle{Proof of \Theorem{Th:convergence}}{.}
\begin{proof}
Given that all the parameters $\theta_v$ are of dimension $L$, that is the dimension of the dictionary $D$, and we have $\Gdims $ nodes, then the total dimension of the parameter $\Theta$ is $LN$.

To facilitate the comparison of \Problem{eq:thetas_cost_function} and the formulation \Expr{eq:LU_opt_problem}, we recall the terms: 
\begin{equation}
\begin{aligned}
h'_{t}  & = (h'_{1,t}, %h'_{2,t},
... h'_{\Gdims,t}) \in \real^{NL}, \ \ 
H'    = 
 \Block(H'_{1,t},...,H'_{N,t}) \in \real^{L \Gdims \times L \Gdims}\!,  \ \ 
H  = 
\Block(H_{1,t},...,H_{N,t}) \in \real^{L \Gdims \times L \Gdims}%\ \ \
\end{aligned}
\end{equation}
Then, we identify the functions $Q$ and $R$ as:
\begin{equation}
\begin{aligned}
Q(\Theta) &=  \frac{1}{\Gdims}  \left( (1-\alpha) \frac{\Theta^\top  H_t \Theta}{2}  + \alpha \frac{\Theta^\top  H'_t \Theta}{2} -   {h'_{t}}^{\top} \Theta \right) +  \frac{\lambda}{2} \Theta^\top (\Lpc \otimes \id_L) \Theta \\
&=  \frac{1}{\Gdims} \sum_{v \in V }  \left( (1-\alpha) \frac{\theta_v^\top  H_{v,t} \theta_v}{2} + \alpha  \frac{\theta_v^\top  H'_{v,t} \theta_v}{2} -  h'^\top_{v,t}  \theta_v \right)  
 + \frac{\lambda}{4} \sum_{u,v \in V}  W_{uv} \norm{\theta_u-\theta_v}^2, \\
R(\theta) &=  \frac{\lambda \gamma}{2} \sum_{v \in V } R_v(\theta_v)  =
\frac{\lambda \gamma}{2} \sum_{v \in V } \norm{\theta_v}^2.
\end{aligned}
\end{equation}
Given this notation, it is easy to verify that updating scheme of \Eq{eq:CCBGD_update} takes the form of \Eq{eq:update_2}.

It is clear that, by definition, that $R(\Theta)$ and $R_v(\theta_v)$ are stronger convex functions of modulus $\lambda \gamma$. Then \Assumption{ass_a:Lipschitz_r} is satisfied. 

Second, the full gradient of $Q(\cdot)$ can be written as:
\begin{equation}
    \Delta Q(\Theta)=   \left( \frac{1-\alpha}{\Gdims} H_t  + \frac{\alpha}{\Gdims} H'_t +  \lambda (\Lpc \otimes \id_{L}) \right) \Theta - \frac{1}{\Gdims} h'_t,
\end{equation}
which is Lipschitz-continuous with constant
%
%\begin{equation}
$M=\eigmax{ \frac{1-\alpha}{\Gdims} H_t  + \frac{\alpha}{\Gdims} H'_t +  \lambda (\Lpc \otimes \id_{L})}$.
%\end{equation}
%
From the node-level expression is easy to derive the partial derivative of $Q(\cdot)$:
\begin{equation}
\begin{aligned}
    \Delta_{\theta_v} Q(\Theta) & = \left(\frac{ (1-\alpha) H_{v,t} + \alpha H'_{v,t} }{\Gdims}\right)\theta_v + \lambda \left( d_v  \theta_v - \sum_{u \in \nghood(v)\!\!\!\!\!\!\!\!} W_{uv} \big( \theta_u \one_{u<v}  + \theta_u \one_{u > v} \big) \right)  -\frac{h'_{v,t}}{\Gdims},
\end{aligned}
\end{equation}
where $d_v$ is the degree of node $v$. This means: 
\begin{equation}
\begin{aligned}
 & \norm{\Delta_v Q(\theta'_{u<v},\theta_v,\theta'_{u>v}) - \Delta_v Q(\Theta')}  \leq \norm{ \left( \frac{ (1-\alpha) H_{v,t} + \alpha H'_{v,t}}{\Gdims}
    + \lambda d_v \id_{L}\right) \left(\theta_v-\theta'_v \right)}  \leq M_v \norm{(\theta_v-\theta'_v)},
\end{aligned}
\end{equation}
where $M_v=  \eigmax{ \frac{ (1-\alpha) H_{v,t} + \alpha H'_{v,t} }{\Gdims}
    + \lambda d_v \id_{L}} $.

Then, \Assumption{ass_a:Lipschitz} is satisfied.

With these elements, and by fixing $\eta_v \mydef M_v$, we can apply Theorem A.1
where
\begin{equation}
\begin{aligned}
M & \mydef \eigmax{ \frac{1-\alpha}{\Gdims} H_t  + \frac{\alpha}{\Gdims} H'_t +  \lambda (\Lpc \otimes \id_{L})} , \ \
M_{v}  \mydef \eigmax{ \frac{ (1-\alpha) H_{v,t} + \alpha H'_{v,t} }{\Gdims} + \lambda d_v \id_{L}}, \ \  
M^{\mu}_{\textup{min}}  \mydef \min_{v \in V} M_v + \mu \\
\mu & \mydef  \lambda \gamma.
\end{aligned}
\end{equation}
After substitution, we get the expression given in \Eq{eq:complexity}.
\end{proof}

\section{Hyperparameter selection strategy}\label{appendix:hyper_selection}

In this section, we detail the model selection strategy to identify the best hyperparameters including the hyperparameters of the Kernel and the penalization constants $\gamma,\lambda$. To make it easier to be read, we denote by $\sigma$ the hyperparameters associated with the Kernel of interest.  The full framework is written in \Alg{alg:model_selection}.

\begin{algorithm}
\small
   \caption{\textbf{--} Hyperparameters tunning for \OCKG \!\!\!\!}\label{alg:model_selection}
\begin{algorithmic}[1]
   \STATE \textbf{Input:}  $\X
   ,\mathcal{X'}$: the adjacent set of observations to be used for estimating the relative \likelihood-ratios: 
\hspace*{\algorithmicindent}\hspace*{\algorithmicindent}\quad
%\begin{equation*}
%\begin{aligned}

$\X:=[\X_{1},...,\X_{\Gdims}]=[[x_{1,1},...,x_{1,n}]
,...,[x_{\Gdims,1},...,x_{\Gdims,n}]]$ 

$\X':=[\X'_{1},...,\X'_{\Gdims}]=[[x'_{1,1},...,x'_{1,n}],...,[x'_{\Gdims,1},...,x'_{\Gdims,n}]]$ \\
%\end{aligned}
%\end{equation*}
$\G=(V,E)$: Observed graph structure; \\
$D$: a precomputed dictionary, associated with the chosen kernel, containing $L$ elements; \\
$\Sigma,\Lambda,\Gamma$: parameter grid of parameters to explore for $\sigma,\lambda,\gamma$; \\
%$\#$: parameter grid of parameters to explore for $\sigma,\lambda,\gamma$; \\
$R$: the number of random splits.\\
 
\STATE \textbf{Output:} $\sigma^*$ the optimal parameter for the kernel $K$, and the two penalization constants $\lambda^*$ and $\gamma^*$.\\
\vspace{1mm}
\hrule
\vspace{1mm}
  \STATE Randomly split $\{1,...,n\}$ into $R$ disjoint subsets $\{I_{r}\}_{r \in \{1,...,R\}}$. Define the subsamples:
	\vspace{-2mm}
\begin{equation*}
\begin{aligned}
&\X_r=[\X_{1,r},...,\X_{\Gdims,r}]=
  [[x_{1,i} , \ i \in I_{r}],
  ..., [x_{\Gdims,i}, \ i \in I_{r}]] \\
& \X_r=[\X'_{1,r},...,\X'_{\Gdims,r}]=
  [[x'_{1,i} , \ i \in I_{r}],
  ...,  [x'_{\Gdims,i}, \ i \in I_{r}]]
\end{aligned}
\vspace{-2mm}
\end{equation*}
  \FOR{ each $\sigma \in \Sigma$}
  \FOR{each $(\lambda,\gamma) \in \Lambda \times \Gamma$}
  \FOR{each data subset $r=1,...,R$} 
	\STATE Let $\X'_{\text{train}} = \X' \backslash \X'_r$, \,$\X'_{\text{test}} = \X'_r$, and $\X_{\text{train}} = \X \backslash \X_r$, \,$\X_{\text{test}} = \X_r$
  \STATE Compute $h'_{v,\text{train}}(\sigma)=\frac{1}{n}\sum_{x \in \X'_{v,\text{train}}} \phi(x,\sigma)$ and $H'_{v,\text{train}}(\sigma)=\frac{1}{n}\sum_{x \in \X'_{v,\text{train}}} \phi(x,\sigma)\phi(x,\sigma)^T$ 
  
\STATE Compute $H_{v,\text{train}}(\sigma)=\frac{1}{n}\sum_{x \in \X_{v,\text{train}}} \phi(x,\sigma)\phi(x,\sigma)^T$. 
	 
   \STATE Solve
\vspace{-2mm}
\begin{equation*}
\hat{\Theta}(\gamma,\lambda)=
\argmin_{\Theta} \frac{1}{\Gdims} \sum_{v \in V} \ell_v(\theta_v) +  \frac{\lambda}{2} \Theta^\top ([\Lpc+\gamma \id_{\Gdims}] \otimes \id_{L}) \Theta
\vspace{-3mm}
\end{equation*}
where%
\vspace{-2mm}
\begin{equation*}
%\vspace{-2mm}
\ell_v(\theta_v)=(1-\alpha) \frac{\theta_v ^\top  H_{v,\text{train}} \theta_v}{2} + \alpha  \frac{\theta_v ^\top  H'_{v,\text{train}} \theta_v}{2}  -   h'_{v,\text{train}} \theta_v 
\vspace{-2mm}
\end{equation*}
\STATE Compute $h'_{v,\text{test}}(\sigma)=\frac{1}{n}\sum_{x \in \X'_{v,\text{test}}} \phi(x,\sigma)$ and $H'_{v,\text{test}}(\sigma)=\frac{1}{n}\sum_{x \in \X'_{v,\text{test}}} \phi(x,\sigma)\phi(x,\sigma)^T$ 
\STATE Compute $H_{v,\text{test}}(\sigma)=\frac{1}{n}\sum_{x \in \X_{v,\text{test}}} \phi(x,\sigma)\phi(x,\sigma)^T$. 
  \STATE Compute
	\vspace{-3mm}
  \begin{equation*}
  \begin{aligned}
  \ell^{(r)}(\hat{\Theta}(\gamma,\lambda)) & \mydef  \frac{1}{\Gdims} \sum_{v \in V} (1-\alpha) \frac{\hat{\theta}_v ^\top  H_{v,\text{test}} \hat{\theta}_v}{2} + \alpha  \frac{\hat{\theta}_v ^\top  H'_{v,\text{test}} \hat{\theta}_v}{2}  -   h'_{v,\text{test}} \theta_v
  \end{aligned}
	\vspace{-2mm}
  \end{equation*}
  \ENDFOR
	\STATE Compute $\hat{\ell}(\sigma,\lambda,\gamma) \mydef \frac{1}{R} \sum_{r=1}^{R} \ell^{(r)}(\Theta)$ \hfill 
  \ENDFOR
  \ENDFOR
\RETURN $\sigma^*,\lambda^*,\gamma^* \mydef  \argmin_{\sigma,\lambda,\gamma}\hat{\ell}(\sigma,\lambda,\gamma) $
\end{algorithmic}
\end{algorithm}

\section{Further experiments}\label{appendix:further experiment}

In order to enhance comprehension, we present the complete experimental setting, by including also the two experiments described in the main text. The implementation details of the four scenarios remains the same as what described in \Sec{sec:exps}.

\inlinetitle{\textup{I}.~Changes in node clusters}{.} 
We sample a Stochastic Block Model with $4$ clusters, $C1$,...,$C4$, each containing $20$ nodes. The intra- and inter-cluster node connection probability is fixed at $0.5$ and $0.01$, respectively.

\textbf{I.a.}~\emph{Bivariate Gaussian distribution to Gaussian copula with uniform marginals}.  

In this first experiment, all nodes will follow a bivariate Gaussian model with the same covariance matrix and mean vector. Then we pick a cluster $C$ at time $t=2000$. From this moment, nodes of $C$ will generate observations from  a Gaussian copula ($\sim GC$) whose marginals follow uniform distributions ($\sim U(-c,c)$): 
\begin{equation}
\begin{aligned}
&(x,y) \sim N(\mu,\Sigma), \ \ \mu=(0,0), \ \ \Sigma_{x,x}=1,\Sigma_{x,y}=\frac{4}{5}  \ \ \rightarrow 
&  (x,y) \sim GC, \ \ \Sigma_{x,x}=1,
\Sigma_{x,y}=\frac{4}{5}. \\
\end{aligned}
\end{equation}
The parameter $c$ is chosen so the mean vector and covariance matrix before and after the change-point are the same. This particular example is hard as the probabilistic model do not depend on the same set of parameters and the first two moments which are used for basic non-parametric methods are the same. 

\textbf{I.b.}~\emph{Change in the covariance matrix or mean vector of bivariate Gaussian distribution}. %
The observations at each node follow a bivariate Gaussian distribution whose parameters $\mu$ and $\Sigma$ depend on the cluster they belong to.
All of them have variance $1$ in both dimensions. We select two cluster of nodes at random and inject a change-point according to the following schema:

\begin{equation}
\begin{cases}
    \mu=(0,0), \Sigma_{1,2}=\frac{4}{5} & \rightarrow \ \  \mu=(0,0), \Sigma_{1,2}=\frac{-4}{5} \ \ \text{if } C1\\
    \mu=(0,0), \Sigma_{1,2}=\frac{4}{5} & \rightarrow \ \  \mu=(0,0), \Sigma_{1,2}=0 \ \ \ \ \text{if } C2\\
    \mu=(0,0), \Sigma_{1,2}=\frac{-4}{5}\!\!\!\! & \rightarrow \ \ \mu=(0,0), \Sigma_{1,2}=0 \ \ \ \ \text{if } C3\\
    \mu=(0,0), \Sigma_{1,2}=\frac{4}{5} & \rightarrow \ \  \mu=(1,1), \Sigma_{1,2}=\frac{4}{5} \ \ \ \ \text{if } C4.\\
\end{cases}
\end{equation}
The set of observations is of length 1000 and the change-point is observed at time $500$. Notice that the difficulty of the detection task will depend on the selected clusters.

\inlinetitle{\textup{II.}~Changes in set of connected nodes}{.} 
In this set of experiments, we sample a Barabási-Albert model with $100$ nodes. This generative model will start with a node at the first iteration, then a node will appear at each iteration $i$ and connect with one of the nodes present at $i-1$ with a probability proportional to their degrees. 

For each instance of the experiments, we will generate $C$ by selecting a node at random with probability proportional to its degree and all the nodes that are at a distance of $4$ in the graph. These nodes will suffer a change in the probability model generating its associated stream. These transitions are described bellow: 

\textbf{II.a.}~\emph{Shift in the mean on one of the cluster components.}  The streams observed at each of the components are drawn from a different 3d Gaussian distribution, before and after the change-point at time $\tau=1000$: 
%%%
\begin{equation}
\begin{aligned}
x_{v} &\sim N(\mu,\Sigma), \mu=\zeros{3}, \Sigma_{i,i}=1,\Sigma_{1,2}=\frac{4}{5},\Sigma_{3,1}=0 \ \ \rightarrow   \ \
x_{v} \sim N(\mu,\Sigma), \mu=(1,0,0), \Sigma_{i,i}=1,\Sigma_{1,2}=\frac{4}{5},\Sigma_{3,1}=0.\!\!\!\!
\end{aligned}
\end{equation}
%%%
\textbf{II.b.}~\emph{Change in probability law.} %
All the nodes will generate observations from standard normal distributions. Then, at time $\tau=2000$, nodes who are in $C$ will start to follow a centralized uniform distributions with unit variance: 
\begin{equation}
x_{v} \sim N(0,1) \ \ \rightarrow \ \ x_{v} \sim U(-\sqrt{3},\sqrt{3}) \ \ \text{if } v \in C. 
\end{equation}

\begin{table}[t!]
\centering
\makebox[\linewidth][c]{%
\scalebox{.883}{
\begin{tabular}{c r || r r r }
    \toprule
 \textbf{\ } & & \textbf{Detection}  %($\downarrow$) 
&    %($\downarrow$) 
&   %($\uparrow$)  
\\
\textbf{\ Scenario} & \textbf{Detector} & \textbf{delay (std)}  %($\downarrow$) 
&   \textbf{AUC (std)}  %($\downarrow$) 
&  \textbf{Precision}  %($\uparrow$)  
\\
    \hline\hline
 %\emph{Synthetic}:
   & \OCKG $\alpha=$ 0.1 & 126.26 \ \ (11.95)  & \textbf{0.89 (0.05)}  &  \textbf{1.00} \ \   \\
Experiment I.A  &     \ \OCKG $\alpha=$ 0.5 & 129.67 \ \ (11.37) &  0.85 (0.06) &  0.98 \ \ \\
$n=$125 &      \  Pool $\alpha=$ 0.1  &  123.72 \ \ (24.08) & 0.82 (0.05)  & 0.58   \ \  \\
    &      \  Pool $\alpha=$ 0.5    & 131.90 \ \ (21.02)  & 0.80 (0.05)  & 0.80  \ \ \\
       &      \  Nougat   & 146.50 \ \ (70.74)  & 0.56 (0.22)  &  0.12   \ \    \\
\hline 
 %\emph{Synthetic}:
   & \OCKG $\alpha=$ 0.1 & 252.72 \ \ (14.82) & \textbf{0.93 (0.03)}  & \textbf{1.00} \ \    \\
Experiment I.A  &     \ \OCKG $\alpha=$ 0.5 & 251.31 \ \ (21.82) & 0.89 (0.04) & 0.98 \ \   \\
$n=$250 &      \  Pool $\alpha=$ 0.1 & 249.54 \ \ (25.20)  & 0.86 (0.04)  & 0.92 \ \  \\
    &      \  Pool $\alpha=$ 0.5  & 245.32 \ \ (22.20) & 0.87 (0.04) & 0.94   \ \  \\
       & \  Nougat &   273.50 \ \  (96.59) & 0.67 (0.19)  & 0.20  \ \  \\
\hline
 %\emph{Synthetic}:
   & \ \OCKG $\alpha=$ 0.1  & 502.70  \ \ \ \ (6.49)  & \textbf{0.99 (0.00)}   & \textbf{1.00} \ \  \\
Experiment I.A  &     \ \OCKG $\alpha=$ 0.5 & 500.84  \ \ \ \ (5.15) & \textbf{0.99 (0.00)}  & \textbf{1.00}  \ \ \\
$n=$ 500 &      \  Pool $\alpha=$ 0.1 & 506.20  \ \   (18.31)  & \textbf{0.99 (0.00)} & \textbf{1.00}  \ \ \\
    &      \  Pool $\alpha=$ 0.5 & 501.90 \ \ \ \ (7.86)   &  \textbf{0.99 (0.00)}  &  \textbf{1.00}  \ \ \\
       &      \  Nougat   &  576.86 (129.27) &  0.66 (0.20) & 0.74   \ \  \\
\hline
 %\emph{Synthetic}:
   & \OCKG $\alpha=$ 0.1 & 25.04  \ \ \ \ (0.20) & \textbf{0.99 (0.02)} & \textbf{1.00}  \ \  \\
Experiment I.B &     \ \OCKG $\alpha=$ 0.5 &  25.06   \ \ \ \ (0.31) & 0.98 (0.02) & \textbf{1.00} \ \  \\
$n=$25 &      \  Pool $\alpha=$ 0.1   & 25.12  \ \ \ \ (0.48) & 0.91 (0.05)  &  \textbf{1.00} \ \  \\
    &      \  Pool $\alpha=$ 0.5   & 25.02   \ \ \ \ (0.24) & 0.96 (0.03) & \textbf{1.00}  \ \ \\
       &      \  Nougat   & 40.68  \ \ \ \ (4.12)  & 0.76 (0.17)   & 1.00  \ \  \\
 \hline
 %\emph{Synthetic}:   
& \OCKG $\alpha=$ 0.1 & 50.18  \ \ \ \ (0.55) & \textbf{1.00 (0.00)} &  \textbf{1.00} \ \ \\
Experiment I.B &     \ \OCKG $\alpha=$ 0.5 & 50.16  \ \ \ \ (0.54) & 0.99 (0.01) &  \textbf{1.00} \ \ \\
$n=$ 50  &      \  Pool $\alpha=$ 0.1   & 50.02  \ \ \ \ (0.32) & 0.93 (0.06)  & \textbf{1.00}  \ \ \\
    &      \  Pool $\alpha=$ 0.5   & 50.00  \ \ \ \ (0.20)  & \textbf{1.00 (0.00)}  &   \textbf{1.00} \ \  \\
       &      \  Nougat   &  67.98  \ \ \ \ (7.63) &  0.78 (0.17) & 0.92 \ \  \\
 \hline
 %\emph{Synthetic}:
   & \OCKG $\alpha=$ 0.1 &  100.12  \ \ \ \ (0.47) & \textbf{1.00 (0.00)} & \textbf{1.00} \ \  \\
Experiment I.B  &     \ \OCKG $\alpha=$ 0.5 & 100.02  \ \ \ \ (0.14) & \textbf{1.00 (0.00)}  &   \textbf{1.00} \ \ \\
$n=$100 &    Pool $\alpha=$ 0.1   &  100.06  \ \ \ \ (0.24) &  \textbf{1.00 (0.01)} & \textbf{1.00} \ \  \\
    &    Pool $\alpha=$ 0.5   &  100.00  \ \ \ \ (0.00)  & \textbf{1.00 (0.00)}   & \textbf{1.00} \ \ \\
  &  Nougat   &  124.23   \ \ (13.50) &  0.88 (0.09) &  1.00 \ \ \\
\hline 
  \hline
 %\emph{Synthetic }:   
& \OCKG $\alpha=$ 0.1  & 25.44  \ \ \ \ (1.96) & \textbf{0.97 (0.02)} & \textbf{1.00}  \ \    \\
Experiment II.A &     \ \OCKG $\alpha=$ 0.5  & 25.06  \ \ \ \ (1.34) & \textbf{0.97 (0.02)}  & 0.96 \ \ \\ 
$n=$25  &      \  Pool $\alpha=$ 0.1  &  24.51  \ \ \ \ (1.68) & 0.91 (0.03)  & 0.82 \ \  \\
    &      \  Pool $\alpha=$ 0.5 & 24.44  \ \ \ \ (2.03)  & 0.93 (0.02) &  0.86 \ \  \\  
       &      \  Nougat  & 34.25  \  \  (13.80)  & 0.64 (0.17)   &   0.08  \ \  \\
 \hline
 %\emph{Synthetic}:
   & \OCKG $\alpha=$ 0.1 & 50.38  \ \ \ \ (1.21)  & \textbf{0.99 (0.01)}  &  \textbf{1.00} \ \  \\
Experiment II.A &     \ \OCKG $\alpha=$ 0.5 & 50.67  \ \ \ \ (1.48) &  0.96 (0.04) &  0.98 \  \  \\ 
$n=$50 &      \  Pool $\alpha=$ 0.1    & 48.55  \ \ \ \ (5.14)  &  0.91 (0.04) & 0.98  \  \   \\
    &      \  Pool $\alpha=$ 0.5  &  49.63  \ \ \ \ (2.38) & \textbf{0.99 (0.01)} &   0.98 \  \ \\   
       &      \  Nougat & 77.84  \ \ (10.24) & 0.75 (0.17)  &  0.76 \  \ \\  
\hline 
 %\emph{Synthetic}:
   & \OCKG $\alpha=$ 0.1  & 100.52  \ \ \ \ (1.25) & 0.99 (0.00)  &  \textbf{1.00} \ \  \\
Experiment II.A &     \ \OCKG $\alpha=$ 0.5 & 100.16  \ \ \ \ (0.64)  & \textbf{1.00 (0.00)}  & \textbf{1.00} \ \  \\
$n=$100 &      \  Pool $\alpha=$ 0.1 & 99.86  \ \ \ \ (1.23)  &  0.99 (0.00) & \textbf{1.00}  \ \  \\  
    &      \  Pool $\alpha=$ 0.5  & 100.38  \ \ \ \ (1.01)  &  0.99 (0.00) & \textbf{1.00} \ \  \\ 
       &      \  Nougat & 127.52  \ \ (13.87)  & 0.77 (0.16)   &  0.88 \ \   \\ 
\hline

%\emph{Synthetic}:
& \OCKG $\alpha=$ 0.1 & 128.17 \ \ \ \  (9.91)  & \textbf{0.86 (0.04)} & 0.94 \ \ \\
Experiment II.B &     \ \OCKG $\alpha=$ 0.5 & 129.63  \ \  (16.66) & \textbf{0.86 (0.04)}  & \textbf{0.96} \ \  \\
$n=$125  &      \  Pool $\alpha=$ 0.1   &  130.05 \ \  (12.48) & 0.79 (0.06)  &   0.88  \ \   \\
    &      \  Pool $\alpha=$ 0.5   & 133.64 \ \  (15.57)  & 0.85 (0.05) & \textbf{0.96} \ \  \\
       &      \  Nougat   & 146.08 \ \  (17.78)  & 0.54 (0.19)  & 0.26  \ \  \\
 \hline
 \hline
 %\emph{Synthetic}:
   & \OCKG $\alpha=$ 0.1 & 249.96 \ \   (20.19) & \textbf{0.96 (0.02)}  &  \textbf{1.00} \ \  \\
    Experiment II.B &     \ \OCKG $\alpha=$ 0.5 & 251.12 \ \  (15.29)  & 0.88 (0.04)  & \textbf{1.00} \ \ \\
$n=$250 &      \  Pool $\alpha=$ 0.1   & 258.51 \ \  (12.18)  & 0.92 (0.03)  & 0.98 \ \  \\
    &      \  Pool $\alpha=$ 0.5   & 254.58 \ \  (19.09)  & 0.88 (0.04) &  1.00 \ \  \\
       &      \  Nougat   & 331.00  (121.15) & 0.49 (0.17)  & 0.34 \ \   \\
\hline 
 %\emph{Synthetic}:
   & \OCKG $\alpha=$ 0.1 & 498.70 \ \ \ \ (8.88) & \textbf{1.00 (0.00)}  & \textbf{1.00} \ \ \\
    Experiment II.B &     \ \OCKG $\alpha=$ 0.5 & 500.00 \ \ \ \ (1.80) & \textbf{1.00 (0.00)} &  \textbf{1.00} \ \ \\
$n=$500 &      \  Pool $\alpha=$ 0.1   & 497.94 \ \  (15.47)  & \textbf{1.00 (0.00)} & \textbf{1.00} \ \ \\
    &      \  Pool $\alpha=$ 0.5   & 500.12  \ \ \ \  (2.28)  & \textbf{1.00 (0.00)}  & \textbf{1.00} \ \ \\
       &      \  Nougat  & 529.86 \ \ (68.53)  &0.39 (0.16)   & 0.98 \ \ \\
\hline 
 \bottomrule
\end{tabular}
}
}
\vspace{-2mm}
\caption{Performance comparison between change-point detectors in several synthetic scenarios for different window sizes ($n$ value). The mean and standard deviation of the score is based on $50$ instances.}\label{tab:synthetic experiments_2}
\end{table}

\begin{figure*}[t]
\centering
%\ \ \ \ \includegraphics[width=0.045\textwidth, viewport=435 235 575 280, clip]{./random.pdf}\\
%\vspace{-3mm}
\centering
\subfigure{\includegraphics[width=0.245\textwidth]{./Experiment2_graph.pdf}
\includegraphics[width=0.245\textwidth]{./Experiment1A_n125.pdf}
\includegraphics[width=0.245\textwidth]{./Experiment1A_n250.pdf}
\includegraphics[width=0.245\textwidth]{./Experiment1A_n500.pdf}}
\vspace{-6mm}
\caption*{\textbf{Experiment I.a}} 
%\vspace{-2mm}
\subfigure{
\includegraphics[width=0.245\textwidth]{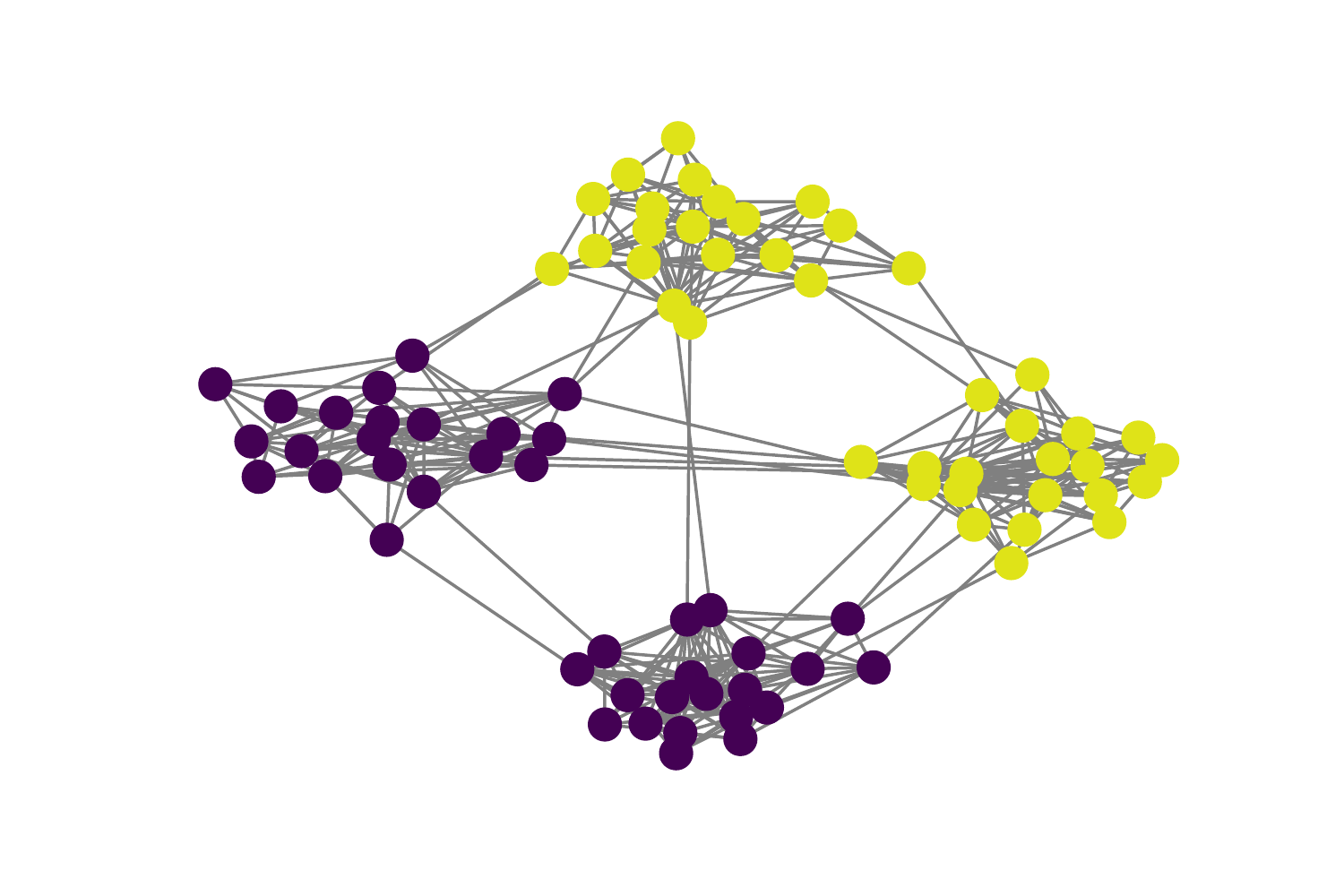}
\includegraphics[width=0.245\textwidth]{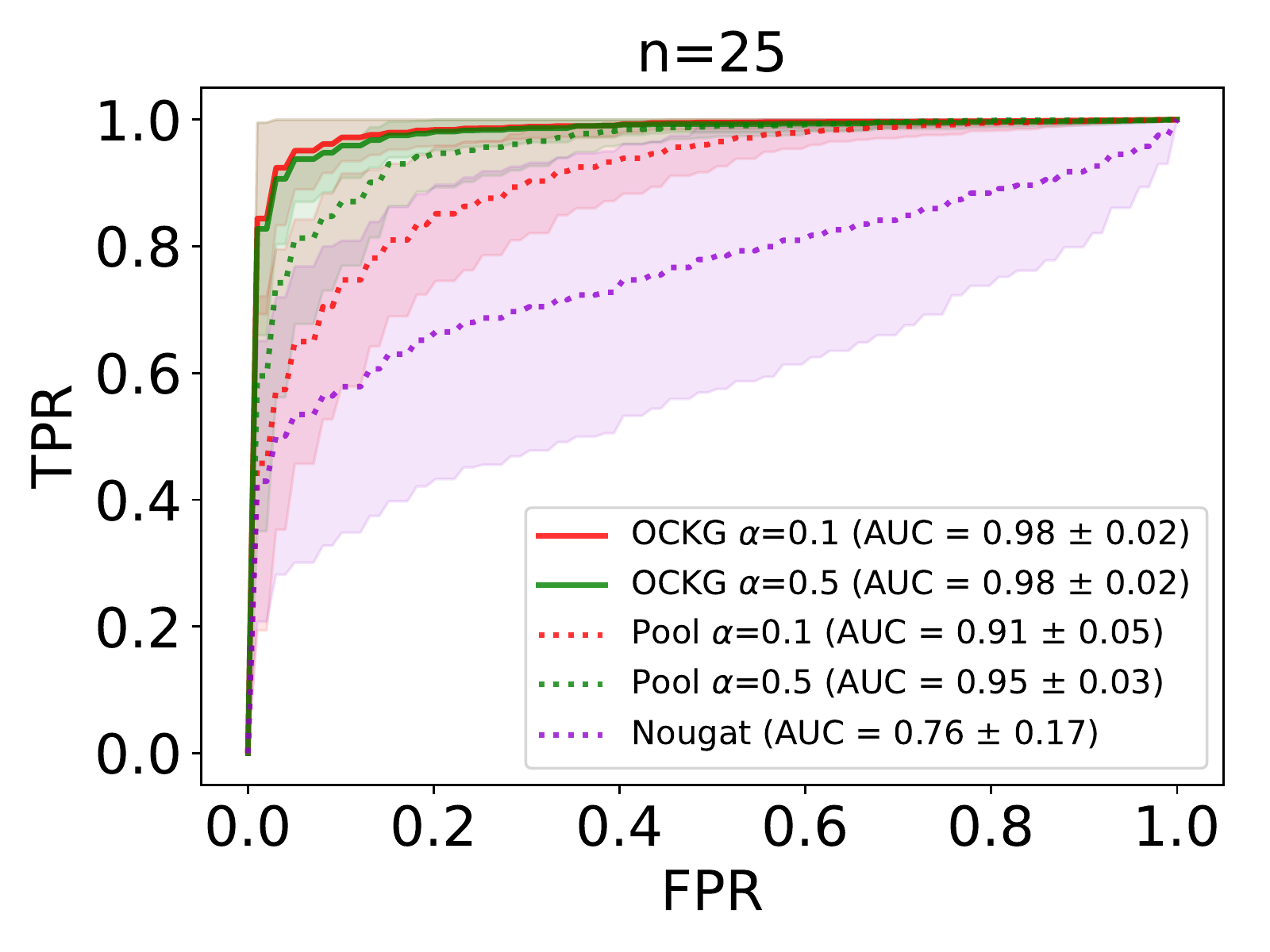}
\includegraphics[width=0.245\textwidth]{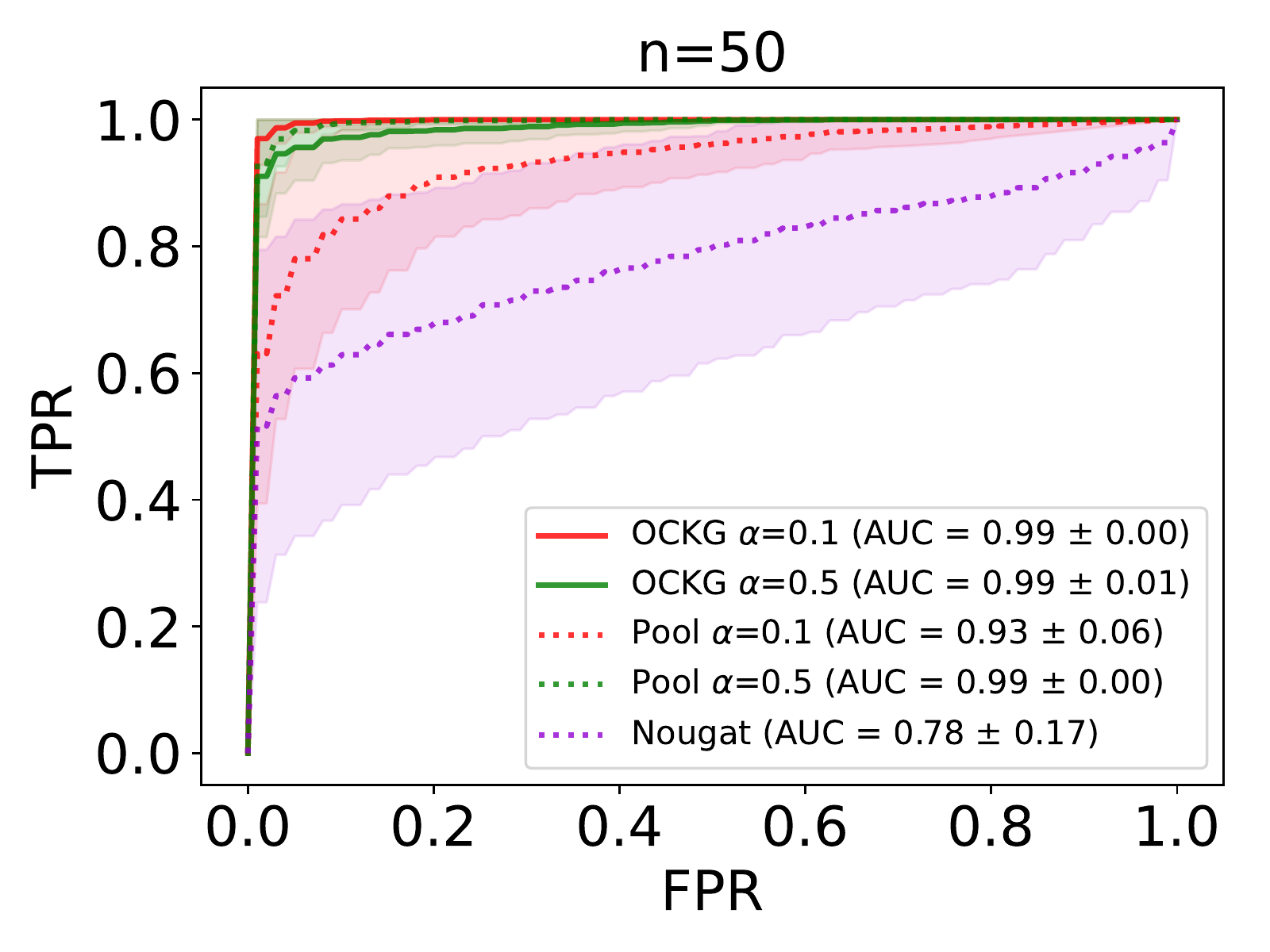}
\includegraphics[width=0.245\textwidth]{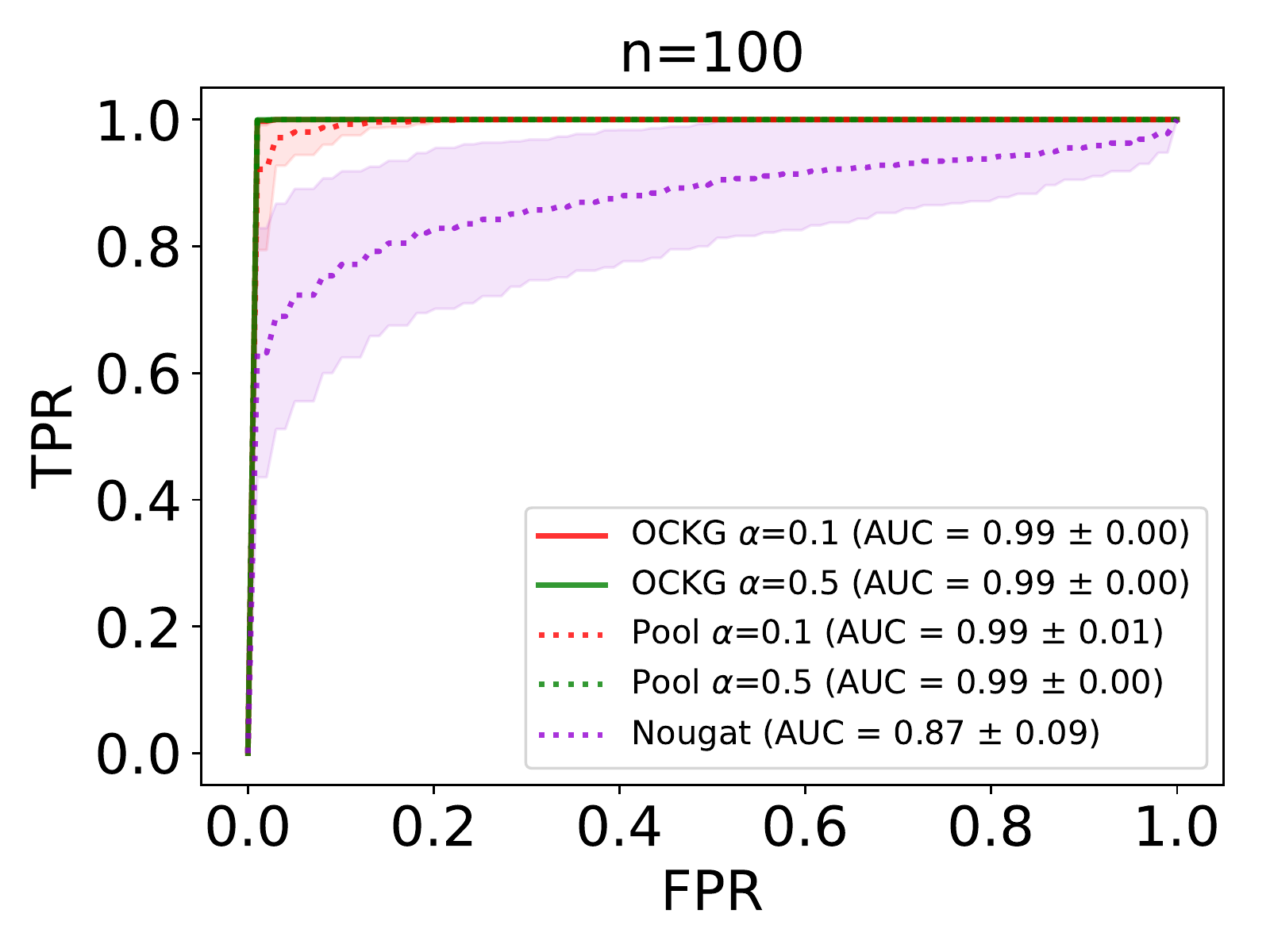}
}
\vspace{-6mm}
\caption*{\textbf{Experiment I.b}} 
%\vspace{-2mm}
\centering
\subfigure{
\includegraphics[width=0.245\textwidth]{./Experiment4_graph.pdf}
\includegraphics[width=0.245\textwidth]{./Experiment2A_n25.pdf}
\includegraphics[width=0.245\textwidth]{./Experiment2A_n50.pdf}
\includegraphics[width=0.245\textwidth]{./Experiment2A_n100.pdf}}
\vspace{-6mm}
\caption*{\textbf{Experiment II.a}} 
%\vspace{-2mm}
\centering
\subfigure{
\includegraphics[width=0.245\textwidth]{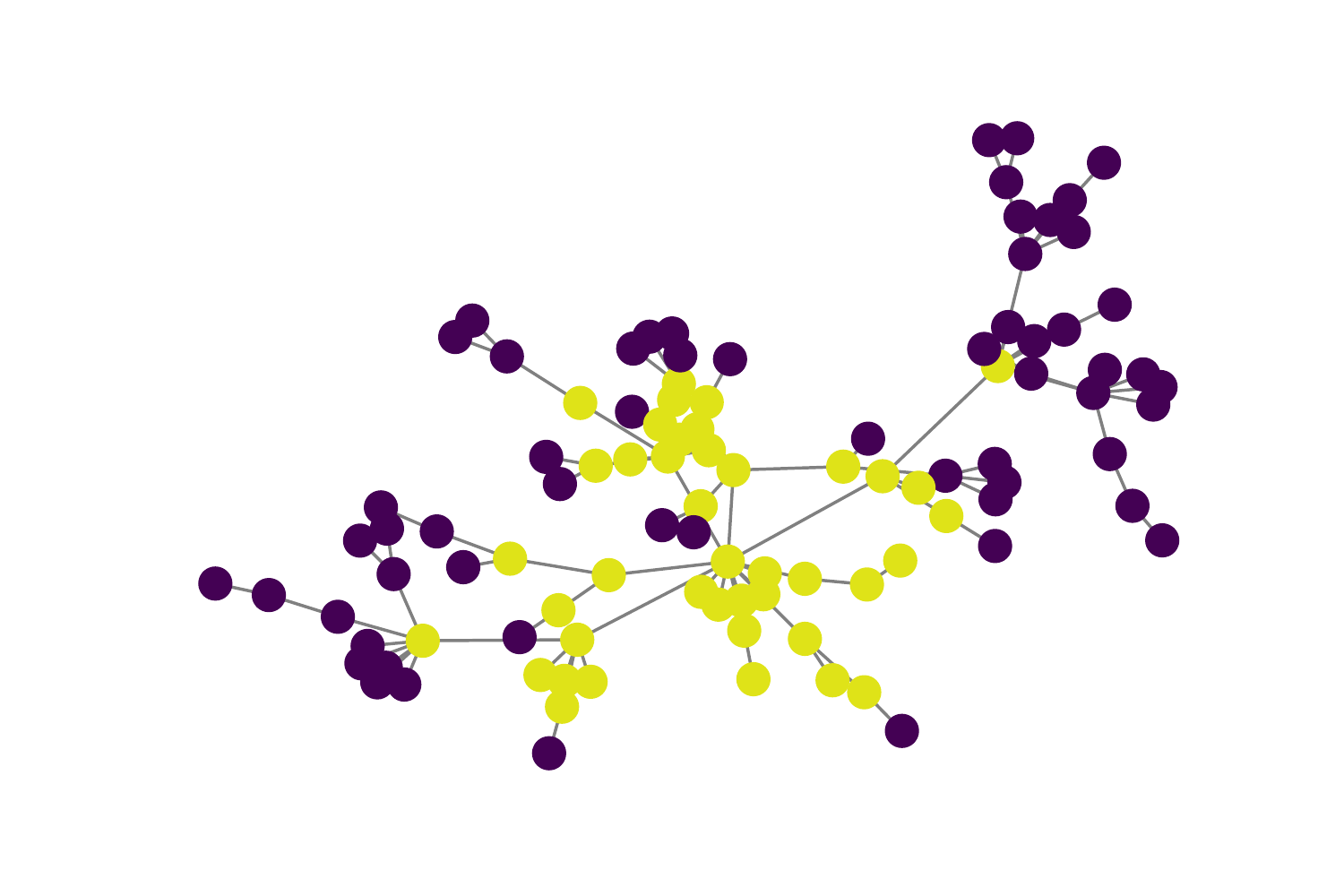}
\includegraphics[width=0.245\textwidth]{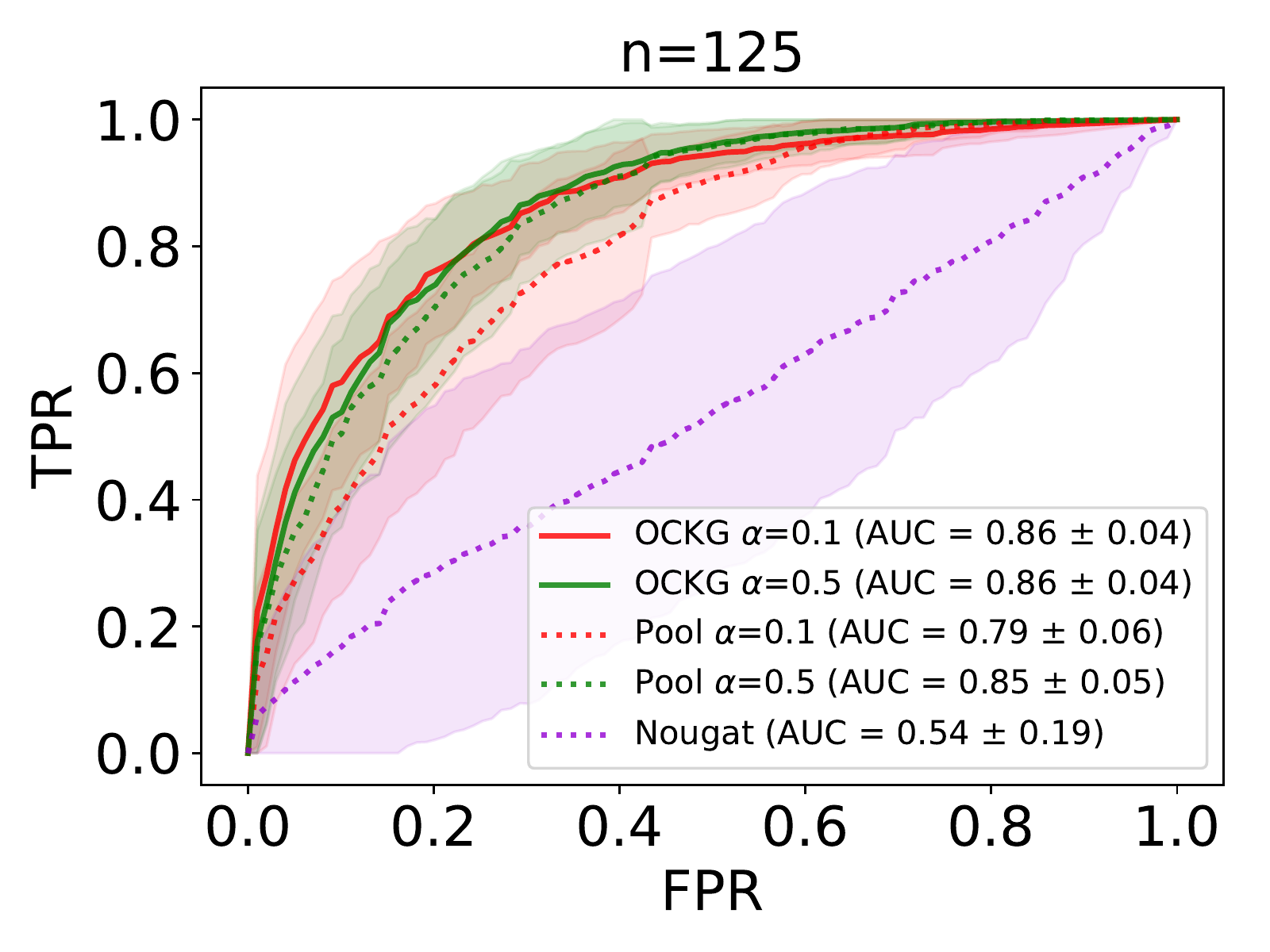}
\includegraphics[width=0.245\textwidth]{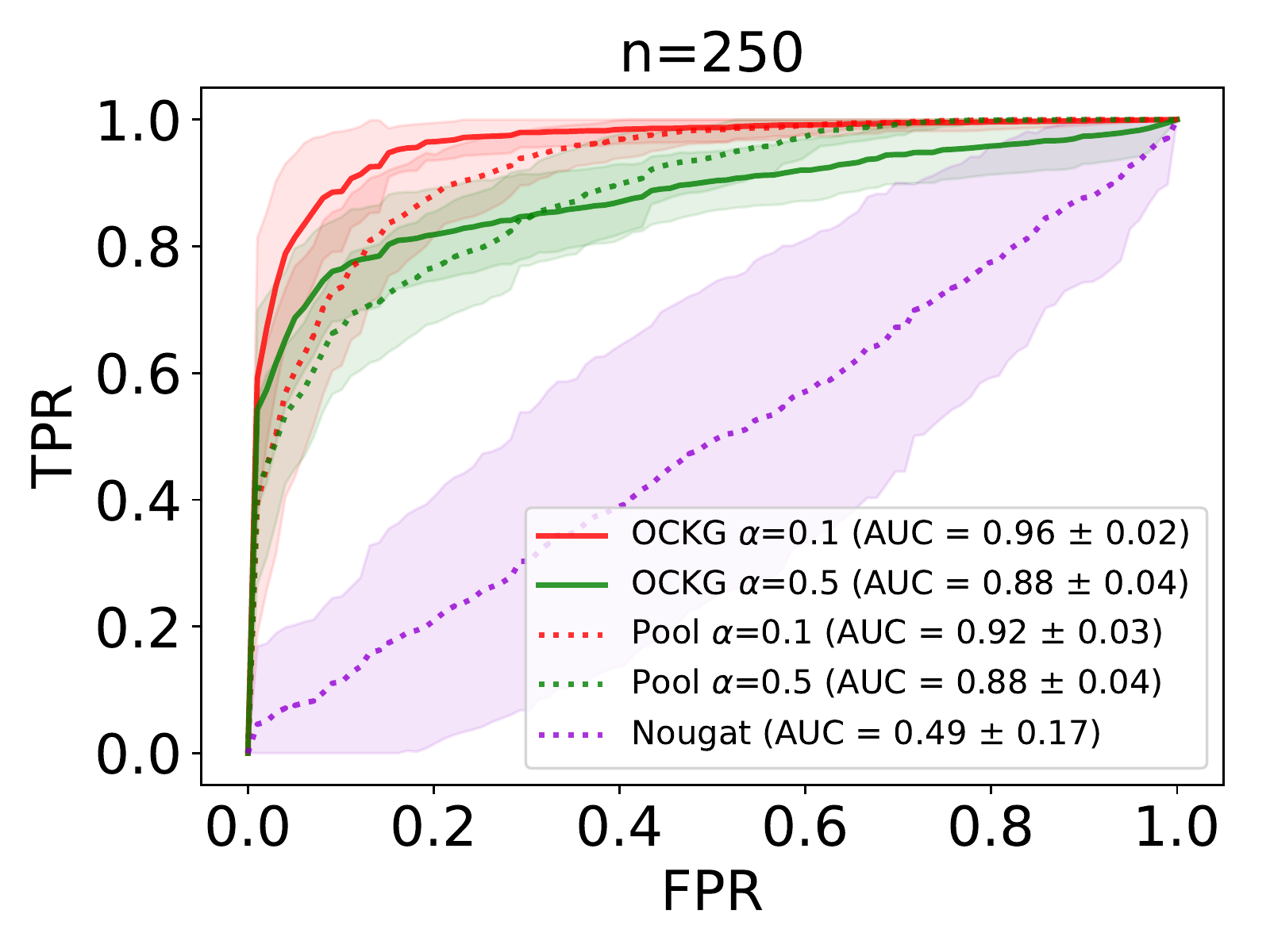}
\includegraphics[width=0.245\textwidth]{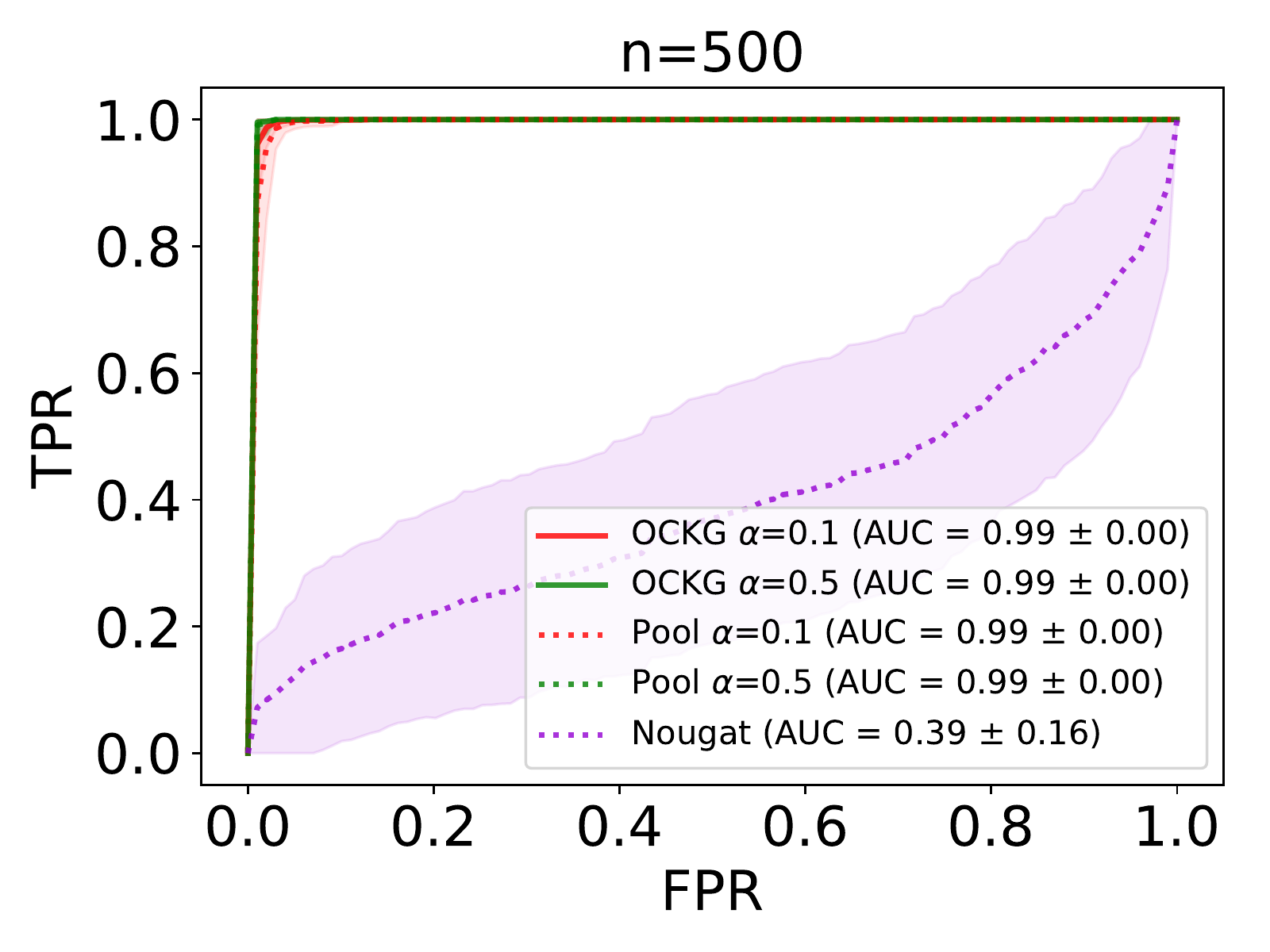}}
\vspace{-6mm}
\caption*{\textbf{Experiment II.b}} 
%\vspace{-2mm}
\caption{Each row presents results for one synthetic scenario. In the first column, an instance of the simulated nodes that suffer a change-point is shown (yellow nodes). The next columns show the mean ROC curves along with their standard deviations for different window sizes $n$. The mean and standard deviations are estimated based on $50$ random instances. %
-- \emph{Experiment I.a}: A bivariate Gaussian distribution changes to a Gaussian copula with uniform marginals. The change affects one cluster of nodes. -- 
\emph{Experiment I.b}: Change in the covariance matrix or mean vector of bivariate Gaussian distribution. The change affects two clusters of nodes. -- \emph{Experiment II.a}: Change in the mean in one dimension of a 3d Gaussian distribution. The change affects a large set of connected nodes. --
\emph{Experiment II.b}: A standardized normal distribution changes to a uniform distribution with the same first two second moments. The change affects a small set of connected nodes.
}\label{fig:results_2}
\end{figure*}

\end{document}